\documentclass[journal]{IEEEtran} 
\usepackage{xcolor}
\usepackage{cite}
\usepackage{hyperref}
\usepackage{amsmath,amsfonts} 
\usepackage{enumerate}  
\usepackage{booktabs}
\usepackage{multirow}
\usepackage{algorithmic}
\usepackage{algorithm}  
\usepackage{graphicx}

\newtheorem{theorem}{Theorem} 
\newtheorem{proposition}{Proposition} 
\newtheorem{definition}{Definition}  
\newtheorem{corollary}{Corollary}

\begin{document}\title{Estimation of Hitting Time by Hitting Probability for Elitist Evolutionary Algorithms}
\author{Jun He~\IEEEmembership{Senior Member,~IEEE},
Siang Yew Chong 
and Xin Yao~\IEEEmembership{Fellow,~IEEE}
 
\thanks{Jun He is with the Department of Computer Science, Nottingham Trent University, Nottingham NG11 8NS, UK. Email: jun.he@ntu.ac.uk.\\
 Siang Yew Chong is with the Department of Computer Science and Engineering, Southern University of Science and Technology, Shenzhen 518055, China, and Centre of Artificial Intelligence and Optimization, School of Computer Science, University of Nottingham Ningbo China. Email: chongsy@sustech.edu.cn. \\
 Xin Yao is with Lingnan University, Tuen Mun, Hong Kong. Email: xinyao@ln.edu.hk.} 
}

\maketitle

\begin{abstract}
Drift analysis is one of the main tools for analyzing the time complexity of evolutionary algorithms. However, it requires manual construction of drift functions to bound hitting time for each specific algorithm and problem.  To address this limitation, linear drift functions were introduced for elitist evolutionary algorithms.  But calculating good linear bound coefficients remains a problem. This paper proposes a new method called drift analysis of hitting probability to compute these coefficients. Each coefficient is interpreted as a bound on the hitting probability of a fitness level, transforming the task of estimating hitting time into estimating hitting probability. A new drift analysis method is then developed to estimate hitting probability, where paths are introduced to handle multimodal fitness landscapes. Explicit expressions are constructed to compute hitting probability, significantly simplifying the estimation process. An advantage of the proposed method is its ability to estimate both the lower and upper bounds of hitting time and to compare the performance of two algorithms in terms of hitting time.  To demonstrate this application, two algorithms for the knapsack problem, each incorporating feasibility rules and greedy repair respectively, are compared. The analysis indicates that neither constraint handling technique consistently outperforms the other.   
\end{abstract}

\begin{IEEEkeywords}
evolutionary algorithms,  hitting time, hitting probability, drift analysis, fitness levels 
\end{IEEEkeywords}

\IEEEpeerreviewmaketitle
\section{Introduction}
\label{secIntroduction}
\IEEEPARstart{H}{itting} time is an important metric to evaluate the performance of evolutionary algorithms (EAs), referring to the minimum number of generations required for an EA to find the optimal solution. Drift analysis is one of the strongest tools used to analyze the hitting time of EAs \cite{doerr2021survey,doerr2012multiplicative} and different drift analysis methods have been developed over the past two decades \cite{kotzing2019first,lengler2020drift,he2001drift,oliveto2011simplified,doerr2013adaptive, mitavskiy2009theoretical,johannsen2010random}. In drift analysis, a drift function is constructed to bound the hitting time, but it is manually tailored for each specific problem \cite{he2001drift}. 

To overcome this limitation, the linear drift function for elitist EAs was proposed \cite{he2023drift}, which combines the strength of drift analysis with the convenience of fitness level partitioning. Given fitness levels $(S_0, \ldots, S_K)$ from high to low, a lower bound on the hitting time from $S_k$ to $S_0$ (where $1 \le k \le K$) is expressed as the following linear function.
\begin{equation}  
\label{equ:Lower-bound}  
 \frac{1}{\displaystyle\max_{X \in S_k} 
p(X,\cup^{k-1}_{j=0}S_j)} +\sum^{k-1}_{\ell=1}\frac{c_{k,\ell}}{\displaystyle\max_{X  \in S_\ell} 
p(X,\cup^{\ell-1}_{j=0}S_j)},
\end{equation}
where the term $p(X,\cup^{\ell-1}_{j=0}S_j)$ represents the transition probability from $X \in S_\ell$ to levels $S_0 \cup \cdots \cup S_{\ell-1}$ and $c_{k,\ell} \in [0,1]$ is a linear coefficient. Similarly, an upper bound on the hitting time from $S_k$ to $S_0$ is expressed as the following linear function.   
\begin{equation} 
\label{equ:Upper-bound} 
 \frac{1}{\displaystyle\min_{X \in S_k} 
p(X,\cup^{k-1}_{j=0}S_j)} +\sum^{k-1}_{\ell=1}\frac{c_{k,\ell}}{\displaystyle\min_{X  \in S_\ell} 
p(X,\cup^{\ell-1}_{j=0}S_j)}.
\end{equation} 

The above drift functions are a family of linear functions used to bound the hitting time for elitist EAs.  Although the calculation of transition probabilities is straightforward, determining the coefficients is more complex.
The primary research question is how to effectively calculate the coefficients for tight linear bounds. Several methods for computing these coefficients have been proposed \cite{wegener2003methods,sudholt2012new,doerr2024lower,he2023drift}. However, more efficient techniques are needed to determine coefficients for tight lower bounds \cite{doerr2024lower}, particularly on multimodal fitness landscapes with shortcuts \cite{he2023drift}. While the linear bound coefficient has been interpreted in terms of visit probability \cite{doerr2024lower}, a general method for computing this visit probability is still lacking. 
 
This paper aims to develop an efficient method for computing the coefficients in linear bounds \eqref{equ:Lower-bound} and \eqref{equ:Upper-bound}. The method makes two significant contributions to the formal development and application of drift analysis. First, it reinterprets a coefficient as a bound on the hitting probability, which is the probability of reaching a fitness level for the first time. Consequently, the task of estimating the hitting time is transformed into estimating hitting probability.

Secondly, drift analysis of hitting probability is introduced to estimate the hitting probability or linear bound coefficients. Although the method is termed ``drift analysis'', it focuses on calculating hitting probabilities but not on hitting times. Hence, it is entirely different from the drift analysis of hitting time \cite{he2001drift}. The method provides a new way to compute linear bound coefficients and introduces new explicit expressions for these coefficients. This greatly simplifies drift analysis because it allows direct estimation of hitting time using explicit formulas. 

Comparing the performance of different EAs is crucial for empirical research. Since the proposed method can estimate both lower and upper bounds of hitting time, it provides a useful tool for theoretically comparing the performance of two EAs in terms of hitting time.   The application is demonstrated through a case study that compares two EAs for the knapsack problem that incorporate feasibility rules \cite{yang2023general,molina2024differential} and solution repair \cite{wang2021set,wang2024novel}, respectively. 
 
This paper is structured as follows. Section \ref{sec:Work} reviews related work. Section \ref{sec:Preliminary} provides preliminary definitions and results. Section \ref{sec:Theory} interprets linear bound coefficients in terms of hitting probabilities. Section \ref{sec:Method} develops a novel drift analysis method for computing hitting probabilities. Section \ref{sec:Comparison} describes the application of comparing two EAs. Finally, Section \ref{sec:Conclusions} concludes the paper. 
 
\section{Related work}
\label{sec:Work} 
 Since the introduction of drift analysis for bounding the hitting time of  EAs \cite{he2001drift}, several variants have been developed, such as simplified drift analysis \cite{oliveto2011simplified}, multiplicative drift analysis \cite{doerr2012multiplicative}, adaptive drift analysis \cite{doerr2013adaptive}, and variable drift analysis \cite{mitavskiy2009theoretical, johannsen2010random}.  A complete review of drift analysis can be found in \cite{kotzing2019first,lengler2020drift}. The main issue in drift analysis is the absence of some universal and explicit expression for the drift function that can be applied to various problems and EAs.

Recently, He and Zhou~\cite{he2023drift} proposed  linear drift functions \eqref{equ:Lower-bound} and \eqref{equ:Upper-bound} designed for elitist EAs.  Based on the coefficients in \eqref{equ:Lower-bound} and \eqref{equ:Upper-bound}, they classified linear drift functions into three categories.
\begin{enumerate}
    \item Type-$c$ time bounds: for all $0< \ell <k\le K$, the coefficients $c_{k,\ell} =c$ are independent on $k$ and $\ell$.
    \item Type-$c_{\ell}$ time bounds: for all $0< \ell <k\le K$, the coefficients $c_{k,\ell} =c_\ell$ depend on $\ell$ but not on $k$.
    \item Type-$c_{k,\ell}$ time bounds: the coefficients $c_{k,\ell}$ depend on both $k$ and $\ell$.
\end{enumerate}

Obviously, Type-$c$ and Type-$c_\ell$ are special cases of the Type-$c_{k,\ell}$ bounds. Wegener~\cite{wegener2003methods} assigned the trivial constants $c_{k,\ell}=0$ for the lower bound and $c_{k,\ell}=1$ for the upper bound. Interestingly, assigning $c_{k,\ell}=1$ provides a tight upper bound for many fitness functions, whereas assigning $c_{k,\ell}=0$ usually leads to a loose lower bound. Several efforts have been made to improve the lower bound. Sudholt~\cite{sudholt2012new} investigated the non-trivial constant $c_{k,\ell}=c$ and used it to derive tight lower bounds for the (1+1) EA on various unimodal functions, including LeadingOnes, OneMax, and long $k$-paths. Sudholt referred to this constant $c$ as viscosity. Doerr and Kötzing~\cite{doerr2024lower} significantly advanced this work by developing a Type-$c_\ell$ lower bound with $c_{k,\ell}=c_\ell$. They applied this to achieve tight lower bounds for the (1+1) EA on LeadingOnes, OneMax, and long $k$-paths jump functions, naming the coefficient $c_\ell$ a visit probability. However, Type-$c$ and Type-$c_\ell$ lower bounds are loose on multimodal fitness landscapes with shortcuts \cite{he2023drift}. 

To address the shortcut issue, He and Zhou \cite{he2023drift} proposed drift analysis with fitness levels and developed the Type-$c_{k,\ell}$ linear bound, though this bound still necessitates recursive computation. Drift analysis with fitness levels has unified existing fitness level methods~\cite{wegener2003methods,sudholt2012new,doerr2024lower} within a single framework. The fitness level method can be viewed as a specific type of drift analysis that employs linear drift functions~\cite{he2023drift}. 

The study of hitting probability has received limited attention in the theory of EAs, with only a few studies available. The term hitting probability has been used in various contexts. He and Yao~\cite{he2002individual} and Chen et al.~\cite{chen2010choosing} used it to denote the probability of reaching an optimum among multiple optima. J{\"a}gersk{\"u}pper \cite{jagerskupper2007algorithmic} used it to describe the probability of a successful step. Yuen and Cheung \cite{yuen2006bounds} referred to it as ``the first pass probability," indicating the probability that the hitting time does not exceed a threshold. K{\"o}tzing~\cite{kotzing2014concentration} also explored this type of ``hitting probability" through negative drift analysis \cite{oliveto2011simplified}. However, none of these studies relate to the explanation of the linear bound coefficients in \eqref{equ:Lower-bound} and \eqref{equ:Upper-bound}. 
 
\section{Preliminaries}
\label{sec:Preliminary}
This section introduces several preliminary definitions and previous results.

\subsection{The Markov Chain for the Elitist EA}
Consider an EA designed to maximize a function \( f(x) \), where \( f(x) \) is defined over a finite set. The EA generates a sequence of solutions \( (X^{[t]})_{t \geq 0} \), where \( X^{[t]} \) represents the solution(s) at generation \( t \). We model the sequence \( (X^{[t]})_{t \geq 0} \) as a Markov chain, following the framework established in \cite{he2003towards, he2016average} . This chain is hereafter referred to as a Markov chain related to the EA. Markov chain theory provides a solid foundation for analyzing the behavior and performance of EAs. Let \( X \in S \) denote a state (a candidate solution), where \( S \) is the state space (all candidate solutions). Let \( S_{\mathrm{opt}} \subseteq S \) denote the subset of optimal solutions. We assume that the chain \( (X^{[t]})_{t \geq 0} \) satisfies three key properties.

\begin{enumerate}
    \item \emph{Convergent (absorbing)}:  Starting from any $X\in S$, the chain can reach (be absorbed into) the optimal set $S_{\mathrm{opt}}$ with probability $1$.
    
    \item \emph{Homogeneous}: The transition probability $p(X,Y)$ from $X$ to $Y$ is independent on $t$.
    \item \emph{Elitist (increasing)}:  Fitness values do not decrease. For any $t \ge 0$, $f(X^{[t+1]})\ge f(X^{[t]})$, where $f(X)$ is the fitness of $X$.
\end{enumerate} 

\subsection{Probability of Transition between Fitness Levels} 
The fitness level method utilizes the transition probabilities between fitness levels. A \emph{fitness level partition $(S_0, \ldots, S_K)$} \cite{wegener2003methods,sudholt2012new,doerr2024lower} is a partition of the state space $S$ into fitness levels according to the fitness from high value to low  such that: 
\begin{enumerate}
    \item  The level $S_0$ is the optimal set $S_{\mathrm{opt}}$.  
    \item For any pair of $X_k \in S_k$ and $X_{k+1} \in S_{k+1}$, the rank order holds: $f(X_k) > f(X_{k+1})$. 
\end{enumerate} 

Thanks to the elitist property, the transition probability from $X_k \in S_k$ to $S_\ell$ (where $0 \le \ell \le K$) satisfies 
\begin{equation}
    p(X_k, S_\ell) = \left\{
    \begin{array}{lll}
      \in [0,1] & \mbox{{if }}\ell \le k,\\
        0 &  \mbox{if } \ell >k.
    \end{array}
    \right.
\end{equation}

Let $[i,j]$ denote the index set $\{i,i+1, \ldots, j-1, j \}$ and $S_{[i,j]}$ denote the union of levels $S_i \cup \cdots \cup S_{j}$.
The transition probability from $X_k$ to $S_{[i,j]}$  is denoted by $p(X_k,S_{[i,j]})$. 
The convergence property implies that  the transition probability \( p(X_k, S_{[0,k-1]}) > 0 \) for any \( k \ge 1 \) and \( X_k \in S_k \).  

The  transition probability  from  $X^{[t]}=X_k$ to $X^{[t+1]}\in S_\ell $ conditional on   $X^{[t+1]} \notin S_k$ is  denoted by
\begin{align}
\label{equ:ConditionalProbability}
r(X_k,S_\ell)    =\left\{
\begin{array}{ll}
\frac{p(X_k, S_\ell)}{p(X_k, S_{[0,k-1]})}   &\mbox{if }\ell < k,\\
0 &\mbox{if } \ell\ge k.
\end{array}
\right.
\end{align}

\subsection{Digraphs and Paths} 
\begin{figure}[ht]
    \centering
\includegraphics[width=0.4\textwidth]{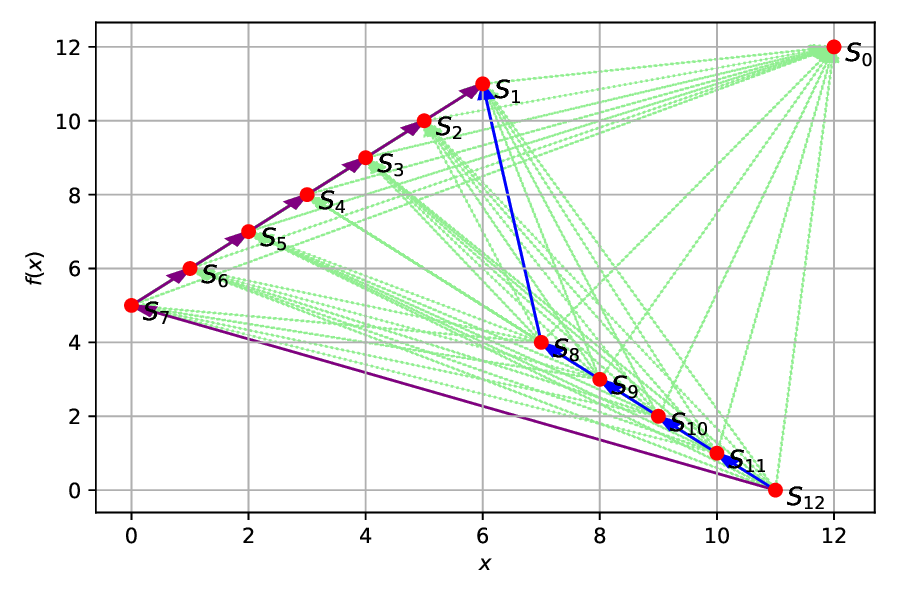}
        \caption{The x-axis represents the state and the y-axis represents the fitness value. Each arc is a transition. Two paths from $S_{12}$ to $S_1$ are highlighted.  }
    \label{fig:Jump}
\end{figure}

Digraphs have been utilized to visualize the behavior of EAs~\cite{chong2019new,chong2019coevolutionary,he2023fast}. In a digraph $(V, A)$, the set $V$ represents the vertices, where vertex $k$ corresponds to level $S_k$. The set $A$ represents the arcs, where arc $(k, \ell)$ indicates the transition from $S_k$ to $S_\ell$, provided that for some $X_k \in S_k$, $p(X_k, S_\ell) > 0$.
Fig.~\ref{fig:Jump} shows an example of a digraph.  

A \emph{path} from $S_k$ to $S_\ell$ is a sequence of distinct vertices $k \to v_{m-1} \to  \cdots \to  v_{1} \to \ell $ and each pair $v_{j+1} \to v_{j} $ is an arc. This path is denoted by $P{[\ell,k]}$. The special path $ k \to k-1 \to \cdots \to \ell+1 \to \ell$ is abbreviated as $[\ell, k]$, which is the same as the index set. Fig.~\ref{fig:Jump} shows two paths from $S_{12}$ to $S_1$. 
Several sub-paths of the path $P{[\ell,k]}$ are represented as follows.
\begin{itemize}  
    \item $P{[\ell,k)}=P{[\ell,k]} \setminus \{k\}$  without vertex $k$.
    \item   $P{(\ell,k]}=P{[\ell,k]} \setminus \{\ell\}
$ without vertex $\ell$. 
    \item $P{(\ell, k)}= P{[\ell,k]} \setminus \{k, \ell\}
$ without vertices $k$ and $\ell$.
\end{itemize}



\subsection{Hitting Time and Hitting Probability} 
Given the Markov chain $(X^{[t]})_{t \ge 0}$ associated with an EA and a fitness level partition $(S_0, \cdots, S_K)$, we introduce the concepts of hitting time and hitting probability based on the textbook \cite{norris1998markov}. 
 
\begin{definition} Assume that the Markov chain $(X^{[t]})_{t \ge 0}$ starts from $X_k \in S_k$, the first \emph{hitting time} from $X_k$ to $S_\ell$  (where $0\le \ell, k \le K$) is 
\[
    \tau(X_k,S_\ell)=\inf \{ t:  X^{[t]} \in S_\ell \}.\]
    The \emph{mean hitting time} $m(X_k,S_\ell)$  denotes the expected value  of $\tau(X_k,S_\ell)$. The \emph{hitting probability} to hit the set $S_\ell$ is 
\[
h(X_k,S_\ell)= \Pr(  \tau(X_k,S_\ell) <\infty).
\] 
The \emph{hitting probability} to hit a specific state $Y_\ell$ in $S_\ell$ is  
\[
h(X_k,Y_\ell)= \Pr(  \tau(X_k,S_\ell)<+\infty \mbox{ and } X^{[\tau(X_k,S_\ell)]} =Y_\ell ).
\]
\end{definition}

The following proposition adapts \cite[Theorem 1.3.2]{norris1998markov} for the hitting probability $h(X_k,S_\ell)$. It states that the hitting probability from $X_k$ to $S_\ell$ is composed of two parts: (i) transitioning from $X_k$ to an intermediate state $Y_i$ and (ii) transitioning from the intermediate state $Y_i$ to $S_\ell$. 

\begin{proposition}
\label{pro:HittingProbability} Given  a fitness level partition $(S_0, \ldots, S_K)$  and two levels $k,\ell: 0\le \ell<k \le K$,
the hitting probability $h(X_k,S_\ell)$ satisfies
\begin{align*} 
 \left\{   
 \begin{array}{ll}
  h(X_k,S_k) =1.   \\
 h(X_k,S_\ell)=\sum^{k}_{i=\ell}\sum_{Y_i \in S_i} p(X_k,Y_i) h(Y_i,S_\ell).
 \end{array}\right.
\end{align*}
\end{proposition}

The following proposition is a modification of \cite[Theorem 1.3.5]{norris1998markov}, adapted to the mean hitting time \( m(X_k, \overline{S}_k) \), where \( \overline{S}_k = S \setminus S_k \) is the complement of the set \( S_k \). It states that the  the mean hitting time \( m(X_k, \overline{S}_k)\) consists of two components: (i) the transition from \( X_k \) to an intermediate state \( Y_k \) within \( S_k \), and (ii) the subsequent transition from \( Y_k \) to a state outside \( S_k \). The value of 1 is included to account for the initial step. 
\begin{proposition}
\label{pro:HittingTime} Given  a fitness level partition $(S_0, \ldots, S_K)$  and a level $k: 0 <k \le K$,
the mean hitting time $m(X_k,\overline{S}_k)$ satisfies
\begin{equation*} 
         m(X_k,\overline{S_k}) 
         = 1+ \sum_{Y_k \in S_k} p(X_k,Y_k) m(Y_k,\overline{S_{k}}).   
\end{equation*}  
\end{proposition}  

For clarity of notation, Table \ref{tab:prob} presents abbreviations for the minimum and maximum values of transition probabilities, hitting probabilities, and mean hitting times. 

\begin{table}[htp] 
\caption{ Abbreviations for minimum and maximum values}
    \label{tab:prob}
    \centering
    \begin{tabular}{c|c }
    \toprule  
          $ p^{{\scriptscriptstyle\min}}_{S_k,S_\ell} :=\displaystyle\min_{X_k \in S_k}p(X_k,S_{\ell})$ & 
         $ p^{{\scriptscriptstyle\max}}_{S_k,S_\ell}:=\displaystyle\max_{X_k \in S_k}p(X_k,S_{\ell})$  \\\midrule
         $ r^{{\scriptscriptstyle\min}}_{S_k,S_\ell}:=\displaystyle\min_{X_k \in S_k}r(X_k,S_{\ell})$ & 
         $ r^{{\scriptscriptstyle\max}}_{S_k,S_\ell}:=\displaystyle\max_{X_k \in S_k}r(X_k,S_{\ell})$  \\\midrule
         $ h^{{\scriptscriptstyle\min}}_{S_k,S_\ell}:=\displaystyle\min_{X_k \in S_k}h(X_k,S_{\ell})$ &
         $ h^{{\scriptscriptstyle\max}}_{S_k,S_\ell}:=\displaystyle\max_{X_k \in S_k}h(X_k,S_{\ell})$\\\midrule
         $ m^{{\scriptscriptstyle\min}}_{S_k,S_\ell}:=\displaystyle\min_{X_k \in S_k}m(X_k,S_{\ell})$ &
         $ m^{{\scriptscriptstyle\max}}_{S_k,S_\ell}:=\displaystyle\max_{X_k \in S_k}m(X_k,S_{\ell})$\\
         \bottomrule
    \end{tabular}    
\end{table}

\subsection{Lower and Upper Bounds on Hitting Time}
Assuming that the chain $(X^{[t]})_{t \ge 0}$ starts from $X$,  $m(X, S_0)$ is the mean hitting time from $X$ to the optimal set $S_0$. If $d(X) \le m(X, S_0)$, then $d(X)$ is called a lower bound of $m(X, S_0)$. Conversely, if $d(X) \ge m(X, S_0)$, then $d(X)$ is called an upper bound of $m(X,S_0)$.

Asymptotic notations such as $O$, $\Omega$, and $o$ are used to differentiate between tight and loose bounds, as described in the textbook \cite{cormen2022introduction}. Let $n$ represent the dimension of the search space. The mean hitting time $m(X,S_0)$ is a function of $n$. A tight bound $d(X)$ differs from the mean hitting time $m(X,S_0)$ by only a constant factor, meaning $d(X) = O(m(X,S_0))$ for an upper bound and $d(X) = \Omega(m(X,S_0))$ for a lower bound. 

\subsection{Previous Results of Drift Analysis Using Linear Drift Functions}
\label{secDrift} 
The main results of drift analysis using linear drift functions \cite{he2023drift} are summarized in Proposition \ref{pro:LinearBound}. A drift function $d(X)$ is used to approximate the mean hitting time $m(X, S_0)$ to the optimal set $S_0$. 
  
\begin{proposition}  
\label{pro:LinearBound}   Given a fitness level partition $(S_0, \ldots, S_K)$,

(1) Let a drift function $d(X)$ satisfy that for any $X_0 \in S_0$,  $d(X_0) = 0$ and for  $1\le k \le K$ and any $X_k \in S_k$, 
\begin{equation*}   
d(X_k)=\frac{1}{p^{{\scriptscriptstyle\max}}_{S_{k},S_{[0,k-1]}}} +\sum^{k-1}_{\ell=1} \frac{c_{k,\ell} }{ p^{\scriptscriptstyle\max}_{S_{\ell},S_{[0,\ell-1]}}},
\end{equation*} 
where coefficients $c_{k,\ell}$ satisfy $c_{\ell,\ell}=1$ and for $k>\ell$,
\begin{equation}
\label{equCoefficientLower}
    c_{k,\ell} \le \min_{X_k \in S_k}  \sum^{k-1}_{j=\ell}  {r}(X_k, S_j) c_{j,\ell}.
\end{equation}
 Then for $k\ge 1$ and any $X_k \in S_k$, the drift
 \[
\Delta d(X_k)= d(X_k)- \sum^{K}_{i=0} \sum_{Y_i \in S_i} p(X_k, Y_i) d(Y_i)  \le 1, 
 \]
 and 
 the mean hitting time
 $ 
m(X_k,S_0) \ge d(X_k).$ 

(2) Let a drift function $d(X)$ satisfy that for any $X_0 \in S_0$,  $d(X_0) = 0$ and for  $1\le k \le K$ and any $X_k \in S_k$, 
\begin{equation*} 
d(X_k)=\frac{1}{p^{{\scriptscriptstyle\min}}_{S_{k},S_{[0,k-1]}}} +\sum^{k-1}_{\ell=1} \frac{c_{k,\ell} }{ p^{\scriptscriptstyle\min}_{S_{\ell},S_{[0,\ell-1]}}},
\end{equation*} 
where coefficients $c_{k,\ell}$ satisfy where coefficients $c_{k,\ell} \in [0,1]$ satisfy $c_{\ell,\ell}=1$ and for $k>\ell$,
\begin{equation}
\label{equCoefficientUpper}
    c_{k,\ell} \ge \max_{X_k \in S_k}    \sum^{k-1}_{j=\ell}  {r}(X_k, S_j) c_{j,\ell}.
\end{equation}
 Then for $k\ge 1$ and any $X_k \in S_k$, the drift
 \[
\Delta d(X_k)=d(X_k)- \sum^{K}_{i=0} \sum_{Y_i \in S_i} p(X_k, Y_i)d(Y_i)\ge 1, 
 \]
 and
 the mean hitting time
 $ 
m(X_k,S_0) \ge d(X_k).$ 
\end{proposition}

In this paper, we interpret \( c_{k,\ell} \) as the hitting probability  from \( X_k \) to \( S_\ell \) and introduce another drift analysis to estimate this hitting probability.

\subsection{Previous Results in the Fitness Level Method}
The Type-$c$ linear bound is a special cases of the Type-$c_{k,\ell}$ bound by setting $c_{k,\ell}=c$ \cite{he2023drift}. 
Sudholt \cite{sudholt2012new} investigated Type-$c$ time bounds. Given a random initial state $X^{[0]}$, he gave the lower time bound as follows:
    \begin{align*} 
        \sum^K_{k=1} \Pr(X^{[0]} \in S_k)  \left[\frac{1}{p^{{\scriptscriptstyle\max}}_{S_{k},S_{[0,k-1]}}} +\sum^{k-1}_{\ell= 1} \frac{c}{p^{\scriptscriptstyle\max}_{S_{\ell},S_{[0,\ell-1]}}}\right].
    \end{align*} 
    where the coefficient $c$ is calculated as follows: 
    \begin{equation} 
\label{equLowerCoeff-c} 
c \le \min_{k:1 < k\le K} \min_{\ell:1\le \ell<k} \; \min_{X_k: p(X_k, S_{[0,\ell]})>0} \frac{p(X_k, S_{\ell})}{p(X_k, S_{[0,\ell]})}.
\end{equation}  
Sudholt \cite{sudholt2012new} used a different expression, but it is equivalent to \eqref{equLowerCoeff-c}.
The constant $c$ is called viscosity, which is a lower bound on the probability of visiting $S_\ell$ conditional on visiting $S_{[0,\ell]}$. 

The Type-$c_\ell$ linear bound is a special cases of the Type-$c_{k,\ell}$ bound by setting $c_{k,\ell}=c_\ell$ \cite{he2023drift}. Doerr and K\"otzing \cite{doerr2024lower} investigated Type-$c_\ell$ time bounds. They gave the lower time bound as follows:  
\begin{equation*}  
 \sum^{K}_{\ell=1}\frac{c_{ \ell}  }{ p^{\scriptscriptstyle\max}_{S_{\ell},S_{[0,\ell-1]}}}.
\end{equation*} 
where the coefficient $c_\ell$ is a lower bound on the probability of visiting $S_\ell$ at least once. It can be calculated as follows, 
\begin{align*}\left\{\begin{array}{ll}
  c_{\ell} \le \min_{k:\ell < k\le K}  \;   \min_{X_k: p(X_k, S_{[0,\ell]})>0} \frac{p(X_k, S_{\ell})}{p(X_k, S_{[0,\ell]})}.  
  \\   c_\ell \le \frac{\Pr(X^{[0]} \in S_{\ell})}{\Pr(X^{[0]} \in S_{[0,\ell]})}.  
\end{array}
\right.
\end{align*}

\section{Estimate Hitting time by Hitting Probability}
\label{sec:Theory}
This section explains the linear bound coefficient as the hitting probability between two fitness levels.

\subsection{Exact Hitting Time} 
Given a fitness-level partition \( (S_0, \ldots, S_K) \), the hitting time from a state \( X_k \in S_k \) to the optimal set \( S_0 \) corresponds to the cumulative time the chain spends in the non-optimal set \( S_1 \cup \cdots \cup S_k \). This intuition is rigorously captured in the following theorem.

\begin{theorem}
\label{the:MeanHittingTime}
Given  a fitness level partition $(S_0, \ldots, S_K)$  and a fitness level $k: 0<k \le K$, the mean hitting time from $X_k \in S_k$ to the optimal set $S_0$ is equal to
    \begin{equation*}     
        m(X_k,S_0) = \sum^{k}_{\ell=1} \sum_{Y_\ell \in S_\ell}  h(X_k,Y_\ell) \, m(Y_\ell, \overline{S}_\ell) .  
    \end{equation*}
\end{theorem}

\begin{IEEEproof} 
When the chain \( (X^{[t]})_{t \ge 0} \) starts from \( X_k \in S_k \), the probability that it first hits a state \( Y_\ell \in S_\ell \) (for \( \ell > 0 \)) is given by the hitting probability \( h(X_k, Y_\ell) \). Upon reaching \( Y_\ell \), the expected time the chain stays in \( S_\ell \) before transitioning to the set \( \overline{S}_\ell\) is \( m(Y_\ell, \overline{S}_\ell) \) where $\overline{S}_\ell= S_0 \cup \cdots \cup S_{\ell-1} $.

Due to the elitist property, once the chain exits \( S_\ell \), it cannot return. The total expected time that is spent in all non-optimal levels \( S_1 \cup \cdots \cup S_k \) before reaching the absorbing set \( S_0 \) is 
\[
\sum_{\ell=1}^{k} \sum_{Y_\ell \in S_\ell} h(X_k, Y_\ell) \, m(Y_\ell, \overline{S}_\ell).
\]

This cumulative time corresponds exactly to the mean hitting time \( m(X_k, S_0) \), since the event of reaching \( S_0 \) for the first time is equivalent to the event of leaving the non-optimal set \( S_1 \cup \cdots \cup S_k \) for the first time.
\end{IEEEproof}


\subsection{Lower and Upper Bounds on Hitting Time}
Theorem \ref{the:MeanHittingTime} gives the exact hit time.
However, in most cases, the exact hitting time cannot be calculated. Therefore, it is necessary to estimate upper and lower bounds. The following theorem gives linear upper and lower bounds on the hitting time based on the hitting probability.

\begin{theorem}
\label{the:TimeBounds}
Given a fitness level partition $(S_0, \ldots, S_K)$ and a fitness level $k: 0<k \le K$, 

(1) The mean hitting time from $X_k \in S_k$ to the optimal set $S_0$ is lower-bounded by 
    \begin{equation}   
    \label{equ:LinearLowerBound2}  m(X_k,S_0)  \ge
 \sum^{k}_{\ell=1}\frac{ h^{\scriptscriptstyle\min}_{S_{k},S_\ell} }{ p^{\scriptscriptstyle\max}_{S_{\ell},S_{[0,\ell-1]}}} .  
    \end{equation}

(2)  The mean hitting time from $X_k \in S_k$ to the optimal set $S_0$ is upper-bounded by
    \begin{equation}   
    \label{equ:LinearUpperBound2}   
    m(X_k,S_0)   \le
 \sum^{k}_{\ell=1}\frac{ h^{{\scriptscriptstyle\max}}_{S_{k},S_\ell} }{ p^{\scriptscriptstyle\min}_{S_{\ell},S_{[0,\ell-1]}}}.  
    \end{equation}
\end{theorem}

\begin{IEEEproof}  1) In the proof, the notation $X^\sharp_\ell \in S_\ell$ denotes a state such that $m(X^\sharp_\ell,\overline{S_\ell})=m^{\scriptscriptstyle\min}_{S_\ell,\overline{S_\ell}}$. First, we estimate a lower bound on $m(X_\ell,\overline{S_\ell})$.   According to  Proposition~\ref{pro:HittingTime}, we have  
 \begin{align*} 
m^{\scriptscriptstyle\min}_{S_\ell,\overline{S_\ell}}&=  m(X^\sharp_\ell,\overline{S_\ell})= 1+\sum_{Y_\ell \in S_\ell} p(X^\sharp_\ell,Y_\ell) m(Y_\ell,\overline{S_{\ell}}) \\
      & \ge  1+ p(X^\sharp_\ell,S_\ell) m^{\scriptscriptstyle\min}_{S_\ell,\overline{S_\ell}}. 
 \end{align*}  
Then we get
  \begin{align}  
        m^{\scriptscriptstyle\min}_{S_\ell,\overline{S_\ell}}
      \ge \frac{1}{1 - p(X^\sharp_\ell,S_\ell)}  
       =   \frac{1}{  p(X^\sharp_\ell,S_{[0,\ell-1]})}  
       \ge  \frac{1}{p^{\scriptscriptstyle\max}_{S_{\ell},S_{[0,\ell-1]}}}. \label{equ:MeanTimeLowerBound} 
 \end{align} 

Second, we estimate a lower bound on the mean hitting time $m(X_k,S_0)$ to the optimal set. 
According to Theorem~\ref{the:MeanHittingTime}, 
\begin{align*}     
          m(X_k,S_0) 
         &= \sum^{k}_{\ell=1} \sum_{Y_\ell \in S_\ell}  h(X_k,Y_\ell) m(Y_\ell, \overline{S}_\ell)\\
        &\ge    \sum^{k}_{\ell=1}   h^{{\scriptscriptstyle\min}}_{S_k,S_\ell}m^{\scriptscriptstyle\min}_{S_\ell,\overline{S_\ell}}  \\
        &\ge \sum^{k}_{\ell=1}\frac{    h^{{\scriptscriptstyle\min}}_{S_k,S_\ell} }{ p^{\scriptscriptstyle\max}_{S_{\ell},S_{[0,\ell-1]}}} \quad \mbox{(by \eqref{equ:MeanTimeLowerBound})} .  
\end{align*}
Then we get Inequality \eqref{equ:LinearLowerBound2}.

2) The proof is similar to the first part.
\end{IEEEproof}

The above theorems provide an explanation of the linear bound coefficients.  The hitting probability  $ h^{{\scriptscriptstyle\min}}_{S_k,S_\ell}$ is a  lower bound coefficient, and  $ h^{{\scriptscriptstyle\max}}_{S_{k},S_\ell} $ is an upper bound coefficient. Any $ c_{k,\ell} \le  h^{{\scriptscriptstyle\min}}_{S_{k},S_\ell}$ is a lower bound coefficient, while any $   c_{k,\ell}  \ge h^{{\scriptscriptstyle\max}}_{S_{k},S_\ell}$  an upper bound coefficient. Since $ h(X_{k},S_k) =1$, the coefficient $c_{k,k}$ is always assigned to $1$.

 The terms viscosity \cite{sudholt2012new}, visit probability \cite{doerr2024lower}, and coefficient \cite{he2023drift}  fundamentally study the same subject: the probability of visiting a fitness level.  Their distinctions  can be characterized as viscosity for Type-\(c\) time bounds, visit probability for Type-\(c_{\ell}\) bounds, and coefficient for Type-\(c_{k,\ell}\) bounds. We use the term hitting probability as it aligns with the standard terminology found in the textbook \cite{norris1998markov}.

\section{Drift Analysis of Hitting Probability}
 \label{sec:Method}
This section outlines a drift analysis method for estimating hitting probabilities.

\subsection{Drift Function} 
Computing exact values of  hitting probabilities is challenging. In this paper, a new drift analysis method is proposed to estimate their bounds. A drift function $c(X_k, S_\ell)$ is used to approximate the hitting probability $h(X_k, S_\ell)$ from $X_k \in S_k$ to $S_\ell$ for any $k, \ell$. It is designed specifically for elitist EAs based on fitness level partitioning.

\begin{definition}
    Given a fitness level partition $(S_0,\ldots,S_K)$,  a \emph{drift function} $c(X_k, S_\ell)$ (where $X_k \in S_k$ and $0\le k, \ell \le K$) is a function such that 
\begin{equation}
\label{equ:DriftFunction}
    c(X_k, S_\ell):=c_{k,\ell} =\left\{
    \begin{array}{ll}
         0 & \mbox{if }    \ell>k,\\
          1    &  \mbox{if }   \ell=k,\\
           \in [0,1]&  \mbox{if }   \ell < k.
    \end{array}
    \right.
\end{equation}  
\end{definition}

The above drift function makes use of two observations that (i) the hitting probability from $S_k$  to $S_\ell$ (where $\ell>k$) is $0$, and (ii) the hitting probability to the same level  is $1$.

\begin{definition} 
Based on the conditional transition probability, the \emph{conditional drift} from $X_k$ to $S_\ell$  is 
\begin{align}
 \tilde{\Delta} c(X_k, S_\ell)&:=  c_{k,\ell} -\sum^{K}_{i=0} \sum_{Y_i \in S_i}  r(X_k, Y_i)    c_{i, \ell} \nonumber
\\
&=c_{k,\ell} -\sum^{k-1}_{i=\ell} r(X_k, S_i) \, c_{i, \ell}\label{equ:ConditionalDrift},
\end{align}  
\end{definition} 

The drift \eqref{equ:ConditionalDrift}  does not contain the terms $i \le \ell$ and $i\ge k$, since for $i \ge k$, $r(X_k, S_i)=0$  and  for $i< \ell$,  $c_{i, \ell}=0$.

\subsection{Drift Conditions} 
The following theorem provides drift conditions to determine that a drift function is a lower or upper bound on the hitting probability. 

\begin{theorem} 
\label{the:DriftTheorem} Given a fitness level partition $(S_0,\ldots,S_K)$, a drift function \eqref{equ:DriftFunction} and two levels $\ell, k: 1 \le \ell <k \le K$,  

(1) If for  $\ell< j \le  k$ and any $X_j \in S_j$,  the conditional drift 
$\tilde{\Delta} c(X_j, S_\ell)  \le 0$, equivalently,  the coefficient  
\begin{align}\label{equ:LowerCoefficient}
    c_{j,\ell} \le \min_{X_j \in S_j}   \sum^{j-1}_{i=\ell}r (X_j, S_i)    c_{i, \ell} ,
\end{align} 
then for any $\ell< j \le  k$, the coefficient  $c_{j,\ell}\le h^{{\scriptscriptstyle\min}}_{S_j,S_\ell}$. 

(2) If for $\ell<j \le k$ and any $X_j \in S_j$,  the conditional drift 
$\tilde{\Delta} c(X_j, S_\ell)  \ge 0$, equivalently, the coefficient  
\begin{align}\label{equ:UpperCoefficient}
    c_{j,\ell} \ge  \max_{X_j \in S_j} \sum^{j-1}_{i=\ell} r (X_j, S_i)    c_{i, \ell},
\end{align} 
then for any $\ell< j \le  k$, the coefficient  $c_{j,\ell}\ge h^{{\scriptscriptstyle\max}}_{S_j,S_\ell}$.  
\end{theorem}

\begin{IEEEproof}  1)  In the proof, the notation $X^\sharp_j \in S_j$ (where $\ell < j\le k$) denotes the state such that $  h(X^\sharp_j, S_\ell) = h^{{\scriptscriptstyle\min}}_{S_j,S_\ell}.$ 

First, we prove that for $\ell < j\le k$, the state $X^\sharp_j$ satisfies the following inequality \eqref{equ:DriftCondition-hmin}.
\begin{align} h^{{\scriptscriptstyle\min}}_{S_j,S_\ell}  \ge   \sum^{j-1}_{i=\ell} r(X^\sharp_j,S_i) h^{{\scriptscriptstyle\min}}_{S_i,S_\ell}\label{equ:DriftCondition-hmin}.
\end{align}
By Proposition~1, the hitting probability from $X^\sharp_j$ to $ S_\ell$ (where $ \ell <j \le k$) satisfies
\begin{align*}
h^{{\scriptscriptstyle\min}}_{S_j,S_\ell}&=  h(X^\sharp_j,S_\ell) 
  =\sum^j_{i=\ell}\sum_{Y_i \in S_i} p(X^\sharp_j,Y_i) h (Y_i,S_\ell)\\
& \ge  \sum^j_{i=\ell}  p(X^\sharp_j,S_i) h^{{\scriptscriptstyle\min}}_{S_i,S_\ell}.
\end{align*} 
Moving the term $i=j$ from the right-hand side sum to the left, we obtain  inequality \eqref{equ:DriftCondition-hmin}
\begin{align*} h^{{\scriptscriptstyle\min}}_{S_j,S_\ell} \ge  \sum^{j-1}_{i=\ell} \frac{p(X^\sharp_j,S_i)}{1-p(X^\sharp_j,S_j) } h^{{\scriptscriptstyle\min}}_{S_i,S_\ell}  = \sum^{j-1}_{i=\ell} r(X^\sharp_j,S_i) h^{{\scriptscriptstyle\min}}_{S_i,S_\ell}.
\end{align*} 
  
Next, using the inequality \eqref{equ:DriftCondition-hmin}, we prove the first conclusion of the theorem, that is, for $j=\ell+1, \cdots, k$,   $c_{j,\ell}\le h^{{\scriptscriptstyle\min}}_{S_j,S_\ell}$ by induction.  
Since $h^{{\scriptscriptstyle\min}}_{S_\ell, S_\ell}=1$, we get   
\begin{align*} 
      c_{\ell+1,\ell}  &\le r(X^\sharp_{\ell+1}, S_\ell) \quad (\mbox{by  \eqref{equ:LowerCoefficient}})\nonumber\\
      & = r(X^\sharp_{\ell+1},S_\ell) h^{{\scriptscriptstyle\min}}_{S_\ell,S_\ell}  \le  h^{{\scriptscriptstyle\min}}_{S_{\ell+1},S_\ell}.  \quad \mbox{(by \eqref{equ:DriftCondition-hmin} )} 
\end{align*}

We make an inductive assumption that for $i=  \ell+1, \ldots, j$,
\begin{equation}
    \label{equ:suppose-a}
    c_{i,\ell}\le h^{{\scriptscriptstyle\min}}_{S_i,S_\ell}.
\end{equation} 
Recall that  $X^\sharp_{j+1}$ satisfies $ h(X^\sharp_{j+1}, S_\ell) =h^{{\scriptscriptstyle\min}}_{S_{j+1},S_\ell}.$ We get 
\begin{align*}
c_{j+1,\ell} &\le \sum^{j}_{i=\ell} r(X^\sharp_{j+1},S_i) c_{i,\ell}  \quad \mbox{(by \eqref{equ:LowerCoefficient})} \\
 &\le \sum^{j}_{i=\ell} r(X^\sharp_{j+1},S_i)  h^{{\scriptscriptstyle\min}}_{S_i,S_\ell}   \quad \mbox{(by \eqref{equ:suppose-a})}
  \\
 & \le  h^{{\scriptscriptstyle\min}}_{S_{j+1},S_\ell}. \quad \mbox{(by \eqref{equ:DriftCondition-hmin}, replace $j$ by $j+1$)}
\end{align*} 

Therefore, the inequality \( c_{j+1,\ell} \le h^{\scriptscriptstyle\min}_{S_{\ell+1}, S_\ell} \) holds. This completes the inductive step, and hence, by induction, the first conclusion is proven. 

2) The proof is similar to the first part.
\end{IEEEproof}

Interestingly, the recursive expressions \eqref{equ:LowerCoefficient} and \eqref{equ:UpperCoefficient} in the above theorem are identical to that in Proposition~\ref{pro:LinearBound}, despite being derived through entirely different proofs. But Theorem \ref{the:DriftTheorem} makes a new contribution: drift analysis of hitting probability. A lower bound coefficient is a lower bound on the hitting probability, while an upper bound coefficient is an upper bound on the hitting probability. A drift function is used to bound the hitting probability, and the drift condition determines whether the drift function serves as a lower or upper bound.

Type-$c$ and Type-$c_\ell$ bounds \cite{he2023drift} can be rewritten in terms of {necessary and sufficient} drift conditions. A detailed analysis is given in the supplementary material. The drift conditions in Theorem \ref{the:DriftTheorem} are based on pointwise drift. Average drift \cite{he2017average} can be used to handle random initialization. 


\subsection{Direct Calculation and Alternative Calculation}
 The coefficients in Theorem \ref{the:DriftTheorem} are determined recursively. He and Zhou~\cite{he2023drift} gave explicit expressions for calculating the coefficients as follows.

\begin{corollary} Given  a fitness level partition $(S_0, \ldots, S_K)$ and two levels $\ell, k: 1 \le \ell <k \le K$,

(1) Let a drift function \eqref{equ:DriftFunction} (where $c(X_k, S_\ell) = c^{\scriptscriptstyle\min}_{k,\ell}$)  satisfy that $c^{\scriptscriptstyle\min}_{\ell, \ell}=1$ and for $1\le \ell<j \le k$,   
\begin{align*} 
   c^{\scriptscriptstyle\min}_{j,\ell}   
    & =   
        r^{{\scriptscriptstyle\min}}_{S_j, S_{\ell}}    + \sum_{ \ell<j_1<j} r^{{\scriptscriptstyle\min}}_{S_j, S_{j_1}}\, r^{{\scriptscriptstyle\min}}_{S_{j_1}, S_{\ell}}  \\ 
&\quad + \sum_{ \ell<j_1<j_2<j} r^{{\scriptscriptstyle\min}}_{S_j, S_{j_2}}\, 
r^{{\scriptscriptstyle\min}}_{S_{j_2}, S_{j_1}}\, r^{{\scriptscriptstyle\min}}_{S_{j_1}, S_{\ell}} 
+ \cdots 
\end{align*}  
Then $
c^{\scriptscriptstyle\min}_{j,\ell} \leq h^{\scriptscriptstyle \min}_{S_j, S_\ell}.
$

(2) Let a drift function \eqref{equ:DriftFunction} (where $c(X_k, S_\ell) = c^{\scriptscriptstyle\max}_{k,\ell}$)   satisfy that $c^{\scriptscriptstyle\max}_{\ell, \ell}=1$ and for $1\le \ell<j \le k$,   
\begin{align*} 
   c^{\scriptscriptstyle\max}_{j,\ell}   
    & = 
        r^{{\scriptscriptstyle\max}}_{S_j, S_{\ell}}  + \sum_{ \ell<j_1<j} r^{{\scriptscriptstyle\max}}_{S_j, S_{j_1}}\, r^{{\scriptscriptstyle\max}}_{S_{j_1}, S_{\ell}}  \\ 
&\quad +\sum_{ \ell<j_1<j_2<j} \,  r^{{\scriptscriptstyle\max}}_{S_{j}, S_{j_2}}  \, r^{{\scriptscriptstyle\max}}_{S_{j_2}, S_{j_1}}   r^{{\scriptscriptstyle\max}}_{S_{j_1}, S_{\ell}} +\cdots  
\end{align*}  
Then $
c^{\scriptscriptstyle\max}_{j,\ell} \geq h^{\scriptscriptstyle \max}_{S_j, S_\ell}.
$
\end{corollary}

\begin{IEEEproof}
(1)    By applying induction, it is straightforward to confirm that for \( \ell < j \leq k \), the coefficient \( c^{\scriptscriptstyle\min}_{j,\ell} \) satisfies 
\[
c^{\scriptscriptstyle\min}_{j,\ell} = \sum^{j-1}_{i=\ell} r^{\scriptscriptstyle\min}_{S_j, S_i}c^{\scriptscriptstyle\min}_{i, \ell} \leq \sum^{j-1}_{i=\ell} r(X_j, S_i) c^{\scriptscriptstyle\min}_{i, \ell}.
\]
Consequently, according to Theorem \ref{the:DriftTheorem}.(1), for \( \ell < j \leq k \),  we have
$
c^{\scriptscriptstyle\min}_{j,\ell} \leq h^{\scriptscriptstyle \min}_{S_j, S_\ell}.
$

 (2) The proof is similar to the first part.
\end{IEEEproof}

Each term in \(c^{\scriptscriptstyle\min}_{j,\ell}\) or \(c^{\scriptscriptstyle\max}_{j,\ell}\) represents the product of conditional probabilities of reaching \(S_\ell\) along a path originating from \(S_j\). The hitting probability is obtained by summing the conditional probabilities over all paths connecting \(S_j\) to \(S_\ell\).  

In Theorem \ref{the:DriftTheorem}, coefficients are computed recursively in the direction from $\ell+1$ to $k$: $c_{\ell+1, \ell}, c_{\ell+2, \ell}, \ldots, c_{k, \ell}$. They can also be computed recursively in the opposite direction from $k-1$ to $\ell$: $c_{k, k-1}, c_{k, k-2}, \ldots, c_{k, \ell}$, as shown below.
 
\begin{theorem} 
\label{the:DriftTheorem-alt}  
Given a fitness level partition $(S_0,\ldots,S_K)$ and two levels $\ell, k: 1 \le \ell <k \le K$,  

(1) Let a drift function \eqref{equ:DriftFunction} satisfy that   for $1\le \ell<j \le k$,  
\begin{align} 
\label{equ:LowerCoefficient-alt}
{c}_{j,\ell} \le  \sum^{j}_{i=\ell+1} {c}_{j,i}  \,  r^{\scriptscriptstyle\min}_{S_i, S_\ell},
\end{align}  
then  for any  $\ell<j \le k$, the coefficient $c_{j,\ell}\le h^{{\scriptscriptstyle\min}}_{S_j,S_\ell}$.

(2) Let a drift function \eqref{equ:DriftFunction} satisfy that for $1\le \ell<j \le k$, 
\begin{align}  
\label{equ:UpperCoefficient-alt}
 {c}_{j,\ell} \ge  \sum^{j}_{i=\ell+1}c_{j,i}  \,  r^{\scriptscriptstyle\max}_{S_i, S_\ell},
\end{align} 
then for any  $\ell<j \le k$, the coefficient $c_{j,\ell}\ge h^{{\scriptscriptstyle\max}}_{S_j,S_\ell}$. 
\end{theorem}

\begin{IEEEproof} 
(1) 
By applying induction to \eqref{equ:LowerCoefficient-alt}, we can establish that \( c_{j,\ell} \leq c^{\scriptscriptstyle\min}_{j,\ell} \). Since $
c^{\scriptscriptstyle\min}_{j,\ell} \leq h^{\scriptscriptstyle \min}_{S_j, S_\ell}
$, we arrive at the desired conclusion.

(2) The proof is similar to the first part.
\end{IEEEproof}

\subsection{Lower Bound Coefficients Using Paths}  
For multimodal fitness landscapes, there are multiple paths from one fitness level to another. For example, in Fig.~\ref{fig:Jump}, there are $11!$ paths from $S_{12}$ to $S_1$. To calculate a lower bound coefficient $c_{k,\ell}$, it is sufficient to use one path $P[\ell,k]$ from $S_k$ to $S_\ell$, rather than using all paths from $S_k$ to $S_\ell$.  For example, in Fig.~\ref{fig:Jump}, coefficient $c_{12,1}$ can be estimated using a longer path $S_{12} \to \cdots \to S_2 \to S_1$, or a shorter path $S_{12} \to \cdots  \to S_8 \to S_1$.  In this case, two values of $c_{12,1}$ can be generated, however, it suffices to utilize any one of them.

The following theorem uses a path to obtain the lower bound coefficient $c_{k,\ell}$. If vertex $j \in (\ell, k]$ is not on the path $P(\ell,k]$, then we directly assign the coefficient $c_{j,\ell }=0$.

\begin{theorem}
\label{the:LowerBoundPaths} Given  a fitness level partition $(S_0, \ldots, S_K)$ and two levels $\ell, k: 1 \le \ell <k \le K$,  let $P{[\ell,k]}$ be a path from $k$ to $\ell$.  Let a drift function \eqref{equ:DriftFunction} satisfy that  for $j \in (\ell,k] \setminus P{(\ell,k]}$,  $c_{j, \ell}=0$,  and for  $j \in P(\ell ,k]$,  
\begin{equation} 
\label{equ:LowerCoefficient3} 
 c_{j,\ell} \le \min_{X_j \in S_j}   \sum_{i \in P{[\ell,j)}} r(X_j, S_i)    \, c_{i, \ell}. 
\end{equation}  
Then  for $\ell <j \le k$, coefficient $c_{j,\ell}\le h^{{\scriptscriptstyle\min}}_{S_j,S_\ell}$. 
\end{theorem}

\begin{IEEEproof}  For  $j\in (\ell, k] \setminus P{(\ell,k]}$, since $c_{j,\ell}=0$, the conditional drift $\tilde{\Delta} c(X_j, S_\ell)   \le 0.$ For $j \in P{(\ell,k]}$, the conditional drift $\tilde{\Delta} c(X_j, S_\ell)  \le 0$ by \eqref{equ:LowerCoefficient3}.   According to Theorem~\ref{the:DriftTheorem}.(1), we get the conclusion.   
\end{IEEEproof}

To avoid recursive computation in \eqref{equ:LowerCoefficient3}, the following corollary provides a non-recursive formula to compute the coefficients. It is a path-based version of  \cite[Theorem 4]{he2023fast}.

\begin{corollary} 
\label{cor:LowerExplicitExpression}  Given  a fitness level partition $(S_0, \ldots, S_K)$ and two levels $\ell, k: 1 \le \ell <k \le K$,    
let $P{[\ell,k]}$ be a path from $k$ to $\ell$.   
Let a drift function \eqref{equ:DriftFunction} satisfy that for $j \in (\ell,k] \setminus P{(\ell,k]}$,   $c_{j, \ell}=0$, and for $j \in P(\ell ,k]$, 
\begin{align}
\label{equ:LowerExplicitExpression}   
 c_{j,\ell} = & \prod_{i\in P(\ell, j]}     r^{\scriptscriptstyle\min}_{S_i,S_{P[\ell,i)}}, 
 \end{align}  
 then for $\ell <j \le k$, coefficient $c_{j,\ell}\le h^{{\scriptscriptstyle\min}}_{S_j,S_\ell}$.  
\end{corollary} 
\begin{IEEEproof}
From the product \eqref{equ:LowerExplicitExpression},  we get  
\begin{align}
& \min_{i \in P{(\ell,j)}}  c_{i, \ell} =  \prod_{i\in P(\ell, j)}     r^{\scriptscriptstyle\min}_{S_i,S_{P[\ell,i)}} , \nonumber\\
\label{equ:Cor1Condition}
 &   c_{j,\ell}  = r^{\scriptscriptstyle\min}_{S_i,S_{P[\ell,i)}} \min_{i \in P{(\ell,j)}}   c_{i, \ell}. 
\end{align} 

For $j \in P{(\ell,k]}$,    
\begin{align*} 
&\min_{X_j \in S_j}  \sum_{i \in P{[\ell,j)}} r(X_j, S_i)    \, c_{i, \ell} \\
 & \ge    \sum_{i\in {P[\ell,j)}}  r^{\scriptscriptstyle\min}_{S_j, S_i}   \min_{i \in P{(\ell,j)}}  c_{i, \ell}\\ 
 &  =  r^{\scriptscriptstyle\min}_{S_j, S_{P{[\ell,j)}}}  \min_{i \in P{(\ell,j)}}   c_{i, \ell} =   c_{j,\ell} \quad \mbox{(by \eqref{equ:Cor1Condition})}.
\end{align*}   

According to Theorem~\ref{the:LowerBoundPaths}, we get the conclusion.  
\end{IEEEproof}

In \eqref{equ:LowerExplicitExpression}, the transition probability $r^{\scriptscriptstyle\min}_{S_i,S_{P[\ell,i)}}$ corresponds to the transition from $S_i$ to the path $P[\ell,i)$. An intuitive interpretation of this corollary is that the hitting probability \( h^{\scriptscriptstyle\min}_{S_k, S_\ell} \) is lower-bounded by the product of the conditional probabilities of staying on the path \( P(\ell, k) \).

\subsection{Upper Bound Coefficients Using Paths} 
Intuitively, it seems impossible to obtain an upper bound coefficient $c_{k,\ell}$ using a path, since the hitting probability of going from $S_k$ to $S_\ell$ is  not less than that of going from $S_k$ to $S_\ell$ via a path $P{[\ell,k]}$. Counterintuitively, the following theorem establishes an upper bound on the hitting probability using one path. If vertex $i \in [\ell, k]$ is not on the path $P(\ell,k]$, we simply assign the coefficient $c_{i,\ell }=1$.

\begin{theorem}
\label{the:UpperBoundPaths}Given  a fitness level partition $(S_0, \ldots, S_K)$  and two levels $\ell, k: 1 \le \ell <k \le K$,    let $P{[\ell,k]}$ be a path from $k$ to $\ell$.
 Let a drift function \eqref{equ:DriftFunction} satisfy that for  $j \in [\ell,k] \setminus P{(\ell,k]}$, $c_{j, \ell}=1$, and for $j \in P(\ell ,k]$,  
\begin{align}\label{equ:UpperCoefficient3}
     c_{j,\ell} \ge   \max_{X_j \in S_j} \left\{ r(X_j, S_{[\ell, j)\setminus P{(\ell,j)}} )
   + \sum_{i\in P{(\ell,j)}} r(X_j, S_i)   c_{i, \ell}\right\}, 
    \end{align}  
 then for  $\ell <j \le k$, the coefficient $c_{j,\ell}\ge h^{{\scriptscriptstyle\max}}_{S_j,S_\ell}$.  
\end{theorem}

\begin{IEEEproof} For $j \in  (\ell,k] \setminus P{(\ell,k]}$, since $c_{j,\ell}=1$, the conditional drift  $\tilde{\Delta} c(X_j, S_\ell) \ge 0.$ For $ j \in P{(\ell,k]}$, the conditional drift $\tilde{\Delta} c(X_j, S_\ell)  \ge 0$ by \eqref{equ:UpperCoefficient3}.    
According to Theorem~\ref{the:DriftTheorem}.(2), we get the conclusion.  
\end{IEEEproof} 

The coefficient computation in Theorem~\ref{the:UpperBoundPaths} is recursive.   
The following corollary provides a non-recursive formula to compute the coefficients. 

\begin{corollary} 
\label{cor:UpperExplicitExpression2} Given  a fitness level partition $(S_0, \ldots, S_K)$  and two levels $\ell, k: 1 \le \ell <k \le K$,    let $P{[\ell,k]}$ be a path from $k$ to $\ell$.  Let a drift function \eqref{equ:DriftFunction} satisfy that for $j \in (\ell,k] \setminus P{(\ell,k]}$, $c_{j, \ell}=1$, and for $j \in P(\ell ,k]$, 
\begin{align}
\label{equ:UpperExplicitExpression2}   
 c_{j,\ell}  =  \sum_{i\in P(\ell, j]}   r^{\scriptscriptstyle\max}_{S_i,S_{[\ell, i)\setminus P(\ell,i)}}, 
 \end{align}
then for $\ell <j \le k$, the coefficient $c_{j,\ell}\ge h^{{\scriptscriptstyle\max}}_{S_j,S_\ell}$.  
\end{corollary} 
\begin{IEEEproof}  From the sum \eqref{equ:UpperExplicitExpression2}, we get 
\begin{align} 
&\max_{i \in P{(\ell,j)}}  c_{i, \ell}  = \sum_{i\in P(\ell, j)}   r^{\scriptscriptstyle\max}_{S_i,S_{[\ell, i)\setminus P(\ell,i)}},\nonumber\\
\label{equ:Cor2Condition}
&  c_{j,\ell}  = r^{\scriptscriptstyle\max}_{S_j, S_{[\ell, j)\setminus P{(\ell,j)}}} +  \max_{i \in P{(\ell,j)}}  c_{i, \ell}.
\end{align} 
 
For $j \in P{(\ell,k]}$,  
\begin{align*} 
&  \max_{X_j \in S_j} \left\{ r(X_j, S_{[\ell, j)\setminus P{(\ell,j)}} )
   + \sum_{i\in P{(\ell,j)}} r(X_j, S_i)   c_{i, \ell}\right\} \\  
&  \le     r^{\scriptscriptstyle\max}_{S_j, S_{[\ell, j)\setminus P(\ell,j)}} +         r^{\scriptscriptstyle\max}_{S_j, S_{P(\ell,j)}} \max_{i \in P{(\ell,j)}} c_{i, \ell}    \\
&  \le     r^{\scriptscriptstyle\max}_{S_j, S_{[\ell, j)\setminus P(\ell,j)}}  +     \max_{i \in P{(\ell,j)}}  c_{i, \ell}  = c_{j,\ell} \quad \mbox{(by \eqref{equ:Cor2Condition})}.
\end{align*}      
According to Theorem~\ref{the:UpperBoundPaths}, we get the conclusion.  
\end{IEEEproof}

In \eqref{equ:UpperExplicitExpression2}, the conditional probability $r^{\scriptscriptstyle\max}_{S_i,S_{[\ell, i)\setminus P(\ell,i)}}$ corresponds to the transitions from $S_i$ to \([\ell,i) \setminus P(\ell, i) \), the vertices not on the path $P(\ell,j)$.  
An intuitive interpretation of this corollary is that the hitting probability \( h^{\scriptscriptstyle\max}_{S_k, S_\ell} \) is upper-bounded by the sum of the conditional probabilities of leaving the path \( P(\ell, k)\) to \([\ell,k) \setminus P(\ell, k) \) .

\section{Comparison of Two Algorithms}
\label{sec:Comparison}
This section applies the proposed method to a comparative analysis of two EAs for the knapsack problem.

\subsection{Comparison of Two EAs}
In computer experiments, the performance of two EAs is assessed using a benchmark suite that includes both easy and hard problems \cite{mahrach2020comparison}. Similarly, theoretical studies should compare EAs using a benchmark suite. 

In this paper, three instances of the knapsack problem are designed to serve as a benchmark suite for comparison. They represent both easy and hard scenarios. The knapsack problem is chosen because of its NP-complete complexity and its well-established role as a classic problem for explaining EAs \cite{michalewicz2014genetic}.  Unlike common benchmarks such as OneMax and LeadingOnes \cite{sudholt2012new,doerr2024lower}, the knapsack problem is subject to a constraint. Various constraint-handling techniques have been employed in EAs, such as feasibility rules \cite{yang2023general,molina2024differential}, the penalty method \cite{zhou2007runtime}, solution repair \cite{wang2021set,wang2024novel}.  

In this section, we compare an EA employing feasibility rules (algorithm 1)  with another EA employing solution repair (algorithm 2). To evaluate their performance, we examine the ratio of their mean hitting times (speedup), defined as 
\[
\frac{ \text{mean hitting time of algorithm 1}}{ \text{mean hitting time of algorithm 2}}.
\]   

\subsection{The Knapsack Problem} 
The knapsack problem is described as follows. There are $n$ items, each with a specific weight $w_i$ and value $v_i$. The goal is to select a subset of these items to include in the knapsack, ensuring that the total weight does not exceed the capacity of the knapsack $C$ while maximizing the overall value. For the $i$th item, let $b_i = 1$ indicate that the item is included in the backpack, and $b_i = 0$ indicate that the item is not included in the backpack. The knapsack problem can be expressed as a constrained optimization problem. Let $x=(b_1, \ldots, b_n)$.
\begin{align}
\label{equ:Knapsack}
    &\max f(x)= \sum^n_{i=1} v_i  b_i 
    &\mbox{subject to }  \sum^n_{i=1}w_i  b_i \le C.
\end{align}

The first EA is the (1+1) EA using feasibility rules, which is described in Algorithm \ref{alg2}. The (1+1) EA is chosen because it serves as a common baseline in the theoretical analysis of EAs \cite{sudholt2012new,doerr2024lower}. Its purpose is to avoid complex calculations of transition probabilities, allowing a focus on the analysis itself. We assume that the EA  is initialized with an empty knapsack; however, other initialization strategies can also be considered.
According to feasibility rules, an infeasible solution will not be accepted because it is worse than the empty knapsack.
 
\begin{algorithm}[ht]
\caption{The (1+1) EA with Feasibility Rules}
\label{alg2}
\begin{algorithmic}[1] 
\STATE Specify $X^{[0]}=x$ to be the empty knapsack. 
\FOR{$t=1,2,\ldots$}
\STATE  Flip  each bit of $x$ independently with  probability $ {1}/{n}$ to generate a solution $y$.
\IF{both $x$ and $y$ are feasible}
\STATE{Select the one with the larger objective value $f$ as $X^{[t+1]}$}.
\ELSIF{both $x$ and $y$ are infeasible}
\STATE{Select the one with the smaller constraint  violation value  $\sum_i w_i b_i-C$ as $X^{[t+1]}$};
\ELSE
\STATE{Select the feasible one as $X^{[t+1]}$}.
\ENDIF
\ENDFOR
\end{algorithmic}
\end{algorithm} 

The second EA is the (1+1) EA using greedy repair, which is described in Algorithm \ref{alg3}. Greedy repair transforms an infeasible knapsack into a feasible one by removing the item(s) with the smallest value-to-weight ratio.  Therefore, it is sufficient to consider feasible solutions. The fitness function is the objective function $f(x)$.
 
\begin{algorithm}[ht]
\caption{The (1+1) EA with Greedy Repair}
\label{alg3}
\begin{algorithmic}[1] 
\STATE Specify $X^{[0]}=x$ to be the empty knapsack. 
\FOR{$t=1,2,\ldots$}
\STATE  Flip  each bit of $x$ independently with  probability $ {1}/{n}$ to generate a solution $y$.
\WHILE{$y$ is infeasible (weight exceeds capacity)}
\STATE Select an item with the smallest value-to-weight ratio and remove it from the knapsack. 
\ENDWHILE
\IF{$f(y) \ge f(x)$}
\STATE $X^{[t+1]}=y$;
\ELSE
\STATE $X^{[t+1]}=x$.
\ENDIF
\ENDFOR
\end{algorithmic}
\end{algorithm}

Table \ref{tab:knapsack} presents three knapsack problem instances with different optimal solutions. In every instance, there are two high-value, high-weight items and \( n - 2 \) low-value, low-weight items. Item 1 has the highest value-to-weight ratio, exceeding 1, while Item 2 has the lowest ratio, falling below 1. The remaining items all have a value-to-weight ratio equals to 1.

\begin{table*}[ht]
    \centering
    \caption{Knapsack problem instances.}
    \label{tab:knapsack}
    \begin{tabular}{ccccc cccc}
    \toprule
         ID &  item $i$ &  $1$ & $2$ &$3,\ldots, n$ & capacity $C$ &global optimum &local optimum  \\
            \midrule
             \multirow{2}{*}{KP1} &  value $v_i$   & $n-2$ & $n/2-1/3$ & $1$ &   \multirow{2}{*}{$n-2$} & \multirow{2}{*}{$L_{(1, 0;0)}$ and $L_{(0,0;n-2)}$} & \multirow{2}{*}{$L_{(0,1;1)}$} \\ 
           & weight $w_i$ & $ n-2-2/3$ & $n-3$ &$1$ &   \\ 
                       \midrule
             \multirow{2}{*}{KP2} &  value $v_i$   & $ n-2-1/3 $ & $n/2-1/3$ & $1$ &   \multirow{2}{*}{$n-2$} & \multirow{2}{*}{ $L_{(0,0;n-2)}$} & \multirow{2}{*}{$L_{(0,1;1)}$ and $L_{(1,0;0)}$} \\ 
            & weight $w_i$ & $ n-2-2/3$ & $n-3$ & $1$ &   \\ 
            \midrule
            \multirow{2}{*}{KP3}  & value $v_i$   & $n-1 $ & $n/2-1/3$ & $1$ &   \multirow{2}{*}{$n-2$} & \multirow{2}{*}{$L_{(1, 0;0)}$  } & \multirow{2}{*}{$L_{(0,1;1)}$ and $L_{(0,0;n-2)}$}
            \\ 
           & weight $w_i$ & $n-2-2/3$ & $n-3$ & $1$ &   \\  
            \bottomrule
    \end{tabular}
\end{table*}

To facilitate analysis, the fitness levels in these knapsack problem instances are expressed in the following form:
\begin{align*}
&L_{(b_1,b_2; k)} = \{x =(b_1,  \ldots, b_n); k= b_3+\cdots +b_n\}.\\
&L_{(b_1,b_2; [i,j])} = \{x=(b_1,  \ldots, b_n); i\le b_3+\cdots +b_n \le j\}.
\end{align*}
Let ${(b_1,b_2;k)}$ denote a solution in $L_{(b_1,b_2;k)}$ and
$L^+_{(b_1,b_2;k)}$ denote the set of feasible solutions with a fitness value larger than $f{(b_1,b_2;k)}$.

The hitting probability from $(a_1,a_2;i)$ to $L_{(b_1,b_2;k)}$ is denoted as $h_{(a_1,a_2;i),(b_1,b_2;k)}$, and its linear bound coefficient is denoted as $c_{(a_1,a_2;i),(b_1,b_2;k)}$. Similarly, $r_{(a_1,a_2;i),(a_1,a_2;i)^+}$ denotes the conditional probability from $(a_1,a_2;i)$ to $L^+_{(a_1,a_2;i)}$, while $m_{(a_1,a_2;i),(b_1,b_2;k)}$ denotes the mean hitting time from $(a_1,a_2;i)$ to $L_{(b_1,b_2;k)}$.
Using the notation, it is convenient to calculate transition probabilities. For example, consider the (1+1) EA with feasibility rules on Instance KP1 and the transition from $L_{(0,0;0)}$ to $L_{(0,0;n-3)}$. This transition happens if and only if bits $b_1,b_2$ remain unchanged, $n-3$ of the $n-2$ zero-valued bits in $b_3, \ldots, b_n$ flips, and the other bits remain unchanged. Therefore, the transition probability
\[
p_{(0,0;0), (0,0;n-3)}=\left(1-\frac{1}{n}\right)^2 \binom{n-2}{n-3} \left(\frac{1}{n}\right)^{n-3} \left(1-\frac{1}{n}\right).
\]

\subsection{Instance KP1} 
\begin{figure}[ht] 
        \centering
\includegraphics[width=0.4\textwidth]{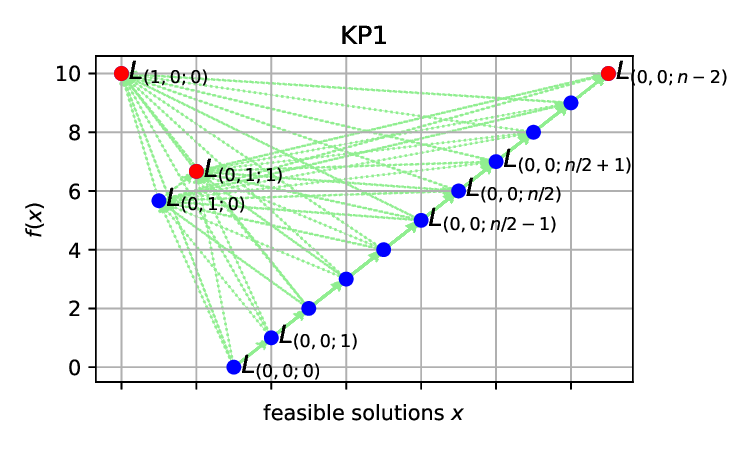} 
        \caption{ The digraph of the two (1+1) EAs on Instance KP1.   Vertices represent fitness levels (feasible solution area). Arcs represent transitions.  
        $n=12$.}
    \label{fig:Knapsack1}
\end{figure}

Fig.~\ref{fig:Knapsack1} shows the digraph of the two (1+1) EAs on Instance KP1. 

\subsubsection{The (1+1) EA Using Feasibility Rules}  
It is sufficient to consider feasible solutions because infeasible solutions are worse than the empty knapsack.  
According to the lower bound \eqref{equ:LinearLowerBound2} in Theorem \ref{the:TimeBounds}, the mean hitting time from the empty knapsack to  the global optimal set $L_{(0,0;n-2)} \cup L_{(1,0;0)}$ is
\begin{align}
    m_{(0,0;0),(0,0;n-2) \cup (1,0;0)}  \ge  \frac{h_{(0,0;0), (0,1;1)}}{p_ {(0,1;1), (0,1;1)^+}}.
\label{instance1-a1} 
\end{align}

Since $
   L^+_{(0,1;1)} =L_{(1,0;0)} \cup L_{(0,0;[n/2+2,n-2])},$
the transition probabilities
\begin{align*} 
  & p_{(0,1;1), (1,0;0)} \le\left(\frac{1}{n}\right)^3,\\
  & p_{(0,1;1), (0,0;[n/2+2,n-2])} \le \frac{1}{n} \binom{n-3}{n/2+1}\left(\frac{1}{n}\right)^{n/2+1}, 
\end{align*}
and then the transition probability 
\begin{align*} 
 p_{(0,1;1), (0,1;1)^+}& \le \left(\frac{1}{n}\right)^3 +\frac{1}{n} \binom{n-3}{n/2+1}\left(\frac{1}{n}\right)^{n/2+1}  \\
 &=O(n^{-3}).
\end{align*}
 
Thus, the mean hitting time 
\begin{align} 
 m_{(0,0;0),(0,0;n-2) \cup (1,0;0)}  \ge \Omega(n^3)\;  h_{(0,0;0), (0,1;1)}. 
      \label{instance1-a3}
\end{align}

We compute the hitting probability $h_{(0,0;0), (0,1;1)}$ using the path $L_{(0, 0;0)} \to  L_{(0,1; 1)}$.  An intuitive observation is that the probability of hitting $L_{(0,1; 1)}$ is $\Omega(\frac{1}{n})$ because of flipping bit $b_2$ and flipping one of bits in $[b_3, b_n]$. Strictly speaking, the hitting probability  
\begin{align*}  
     h_{(0,0;0),(0,1;1)}  & \ge p_{(0,0;0),(0,1;1)}  \\ 
&\ge \frac{1}{n} \binom{n-2}{1} \frac{1}{n} \left(1-\frac{1}{n}\right)^{n-2}  =  \Omega\left(\frac{1}{n}\right).
\end{align*}

Thus, the mean hitting time from the empty knapsack to  the global optimal set $L_{(0,0;n-2)} \cup L_{(1,0;0)}$ is
\begin{align}  \label{instance1-b3}
 m_{(0,0;0),(0,0;n-2) \cup (1,0;0)}  =    \Omega(n^3)   \Omega(n^{-1})     =\Omega(n^2). 
\end{align}  

\subsubsection{The (1+1) EA with Greedy Repair}  
Let $S_0$ denote the global optimal set $L_{(0,0;n-2)} \cup L_{(1,0;0)}$, and $S_1$ be the rest of feasible solutions.
According to the upper bound \eqref{equ:LinearUpperBound2} in Theorem \ref{the:TimeBounds},
the mean hitting time from any $x_1 \in S_1$  to $S_0$ is upper-bounded by
\begin{align*}
\frac{1}{\min_{x_1 \in S_1}p(x_1;S_0)}.
\end{align*}

For any $x_1=(0,b_2;k) \in S_1$, the probability of a mutation from $(0,b_2;k)$ to $(1,*;*)$ is $\frac{1}{n}$ (where $*$ represents an arbitrary value). Since Item 1 has the largest value-to-weight ratio, after greedy repair, only item 1 remains and the solution becomes $(1,0;0)$. So the  probability   $p(x_1;S_0)$ is  at least
$\frac{1}{n}.$ 
Then,  the mean hitting time from the empty knapsack to the global optimum $(1,0;0)$ is
\begin{align}
\label{instance1-c1}
    m_{(0,0;0),(1,0;0)}=O(n).
\end{align}

By comparing equations \eqref{instance1-b3} and \eqref{instance1-c1}, we find that for KP1,  the (1+1) EA with greedy repair is faster than that of the (1+1) EA using feasibility rules by a factor of \(\Omega(n)\). 


\subsection{Instance KP2}
\begin{figure}[ht] 
        \centering
\includegraphics[width=0.4\textwidth]{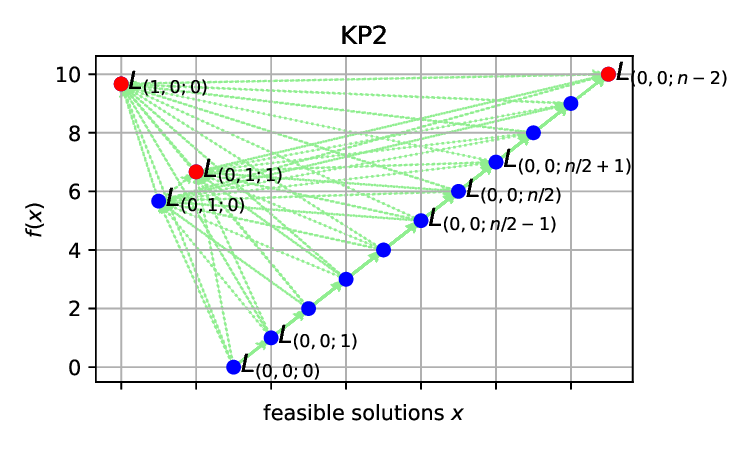}
        \caption{ The digraph of the two (1+1) EAs on Instance KP2.  Vertices represent fitness levels (feasible solution area). Arcs represent transitions.  
        $n=12$.}
    \label{fig:Knapsack2}
\end{figure}

Fig.~\ref{fig:Knapsack2} shows the digraph of the two (1+1) EAs on Instance KP2.

\subsubsection{The (1+1) EA Using Feasibility Rules}   
According to the upper bound \eqref{equ:LinearUpperBound2} in Theorem \ref{the:TimeBounds}, the mean hitting time from the empty knapsack to the global optimum is 
\begin{align}
    &m_{(0,0;0), (0,0;n-2)} \le   \sum^{n-3}_{\ell=0} \frac{1}{p_{(0,0;\ell), (0,0;\ell)^+}}+
  \frac{1}{p_{(0,1;0), (0,1;0)^+}}\nonumber\\
    & \quad +
  \frac{1}{p_{(0,1;1), (0,1;1)^+}}+
  \frac{h_{(0,0;0), (1,0;0)}}{p_{(1,0;0), (1,0;0)^+}}.
\label{instance2-a1} 
\end{align}  

The transition probabilities in \eqref{instance2-a1} are estimated as follows. For $\ell=0, \ldots, n-3$, the transition probabilities 
\begin{align*}
&p_{(0,0;\ell), (0,0;\ell)^+} \ge p_{(0,0;\ell),  (0,0;\ell+1)} \ge  \frac{n-2-\ell}{n} \left(1-\frac{1}{n}\right)^{n-1} . \\
  &p_{(0,1;0), (0,1;0)^+} \ge p_{(0,1;0), (0,1;1)} =  { \frac{n-2}{n}   \left(1-\frac{1}{n}\right)^{n-1}.}
\end{align*}
\begin{align*}
  &p_{(0,1;1), (0,1;1)^+} \ge p_{(0,1;1), (1,0;0)} \ge \left(\frac{1}{n}\right)^3   \left(1-\frac{1}{n}\right)^{n-3}.
\\
&  p_{(1,0;0), (1,0;0)^+} \ge p_{(1,0;0), (0,0;n-2)} \ge   \left(\frac{1}{n}\right)^{n-1}   \left(1-\frac{1}{n}\right) .
\end{align*}

Thus, we get  the mean hitting time 
\begin{align} 
  m_{(0,0;0), (0,0;n-2)} \le          O(n^3)+  \frac{h_{(0,0;0), (1,0;0)}}{(n-1)n^{-n}} . 
      \label{instance2-a4}
\end{align} 
 
Intuitively, the hitting probability $h_{(0,0;0), (1,0;0)}$ is no more than the sum of the conditional transition probabilities from $L_{(0,0;\ell)}$ to $L_{(1,0;0)} \cup L_{(0,1;[0;1])}$ where $\ell =0,\ldots, n-3$. Since these conditional probabilities decreases exponentially fast as $\ell$ increases, the hitting probability $h_{(0,0;0), (1,0;0)}=O(\frac{1}{n})$. We rigorously prove this using Corollary \ref{cor:UpperExplicitExpression2}.

We choose the  path $L_{(0,0;0)} \to L_{(0,0;1)} \to  L_{(0,0;2)} \to \cdots \to L_{(0,0;n-3)} \to L_{(1,0;0)}$ to calculate the hitting probability   $h_{(0,0;0),(1,0;0)}$.  According to Corollary \ref{cor:UpperExplicitExpression2}, the hitting probability  
 \begin{align}  
   & h_{(0,0;0),(1,0;0)} \le  \sum^{n-3}_{\ell=0}  r_{(0,0;\ell), (1,0;0)\cup (0,1;[0,1])} \nonumber
\\&= \sum^{n-3}_{\ell=0}  \frac{p_{(0,0;\ell), (1,0;0) }+p_{(0,0;\ell),  (0,1;0)}+p_{(0,0;\ell),  (0,1;1)}}{p_{(0,0;\ell), (0,0;\ell)^+}}\nonumber 
\\&\le \sum^{n-3}_{\ell=0}  \frac{p_{(0,0;\ell), (1,0;0) }+p_{(0,0;\ell),  (0,1;0)}+p_{(0,0;\ell),  (0,1;1)}}{  p_{(0,0;\ell), (0,0;\ell+1)}} . 
     \label{instance2-a9}
\end{align}

The transition probabilities
\begin{align*}
    &p_{(0,0;\ell), (1,0;0)} \le   \left(\frac{1}{n}\right)^{\ell+1}.  
    \\
    &p_{(0,0;\ell), (0,1;0)}\le 
         \left(\frac{1}{n}\right)^{\ell+1} . \\
    &p_{(0,0;\ell), (0,1;1)} \le \left\{\begin{array}{ll}
         \frac{1}{n} , & \ell\le 1,  \\
        \frac{1}{n}\binom{\ell}{\ell-1} \left(\frac{1}{n}\right)^{\ell} ,& \ell \ge 2.
    \end{array}\right. \\
&p_{(0,0;\ell), (0,0;\ell+1)}   \ge  \frac{n-2-\ell}{n} \left(1-\frac{1}{n}\right)^{n-1}.
\end{align*}

Substituting them into \eqref{instance2-a9}, we get the hitting probability  
\begin{align} h_{(0,0;0),(0,1;0)} &\le O\left(\frac{1}{n}\right)+ \sum^{n-3}_{\ell=2} \frac{e}{n-2-\ell}    \left(\frac{ 2   }{  n^{\ell} }  +  \frac{  \ell}{n^{\ell} }\right)
     \nonumber\\
     &= O\left(\frac{1}{n}\right).
     \label{instance2-a20}
\end{align} 

Inserting \eqref{instance2-a20} into \eqref{instance2-a4}, we get the mean hitting time  $m_{(0,0;0), (0,0;n-2)}$ is upper-bounded by
\begin{align}    
      \label{instance2-a5}  
     m_{(0,0;0), (0,0;n-2)} =   \frac{O(n^{-1})}{(n-1)n^{-n}}.
\end{align}

\subsubsection{The (1+1) EA Using Greedy Repair}  
According to the lower bound \eqref{equ:LinearLowerBound2} in Theorem \ref{the:TimeBounds}, the mean hitting time from the empty knapsack to the global optimum is
\begin{align}
     m_{(0,0;0),(0,0;n-2)} \ge    \frac{h_{(0,0;0),  (1,0;0)}}{p_{ (1,0;0), (1,0;0)^+}}.
\label{instance2-b1} 
\end{align}

Since $L^+_{(1,0;0)} = L_{(0,0;n-2)}$, 
 the transition probability 
 \begin{align}  \label{instance2-b2}
   p_{(1,0;0), (0,0;n-2)} \le \left(1-\frac{1}{n}\right)  \left(\frac{1}{n}\right)^{n-1} .
\end{align} 
Then, the mean hitting time 
\begin{align} 
    m_{(0,0;0),(0,0;n-2)}\ge        \frac{h_{(0,0;0), (1,0;0)}}{ (n-1)n^{-n}}.
      \label{instance2-b3}
\end{align}

Intuitively, the chain follows the path $(0,0;0) \to (0,0;1) \to \cdots \to (0,0;n/2-1)$ with probability $\Omega(1)$. For each vertex on this path, the mutation probability from $(0,0;\ell)$ (where $\ell=0, \ldots, n/2-1$) to $ (1,*;*)$ is $\frac{1}{n}$. After greedy repair, the solution becomes $(1,0;0)$. Thus, the hitting probability $h_{(0,0;0), (1,0;0)}$ is no less than the sum of $\Omega(\frac{1}{n})$ over $\ell=0, \ldots, n/2-1$. Then the hitting probability $h_{(0,0;0), (1,0;0)}=\Omega(1)$. We rigorously prove this using Theorem \ref{the:DriftTheorem-alt}.
 
According to \eqref{equ:LowerCoefficient-alt} in Theorem \ref{the:DriftTheorem-alt}, we have a lower bound on the hitting probability as
\begin{align}
&
   h_{(0,0;0), (1,0;0)} \ge  c_{(0,0;0), (1,0;0)}   \nonumber\\
    &=  r_{(0,0;0), (1,0;0)} +\sum^{n/2-1}_{\ell=1} c_{(0,0;0), (0,0;\ell)} \, r_{(0,0;\ell), (1,0;0)}.
    \label{instance2-b4}
\end{align} 
We omit the terms with \( \ell \ge \frac{n}{2} \) from the summation, as excluding them still yields a valid lower bound. 

The conditional transition probability $r_{(0,0;\ell), (1,0;0)}$ (where $0\le \ell <n/2$) is calculated as follows. The mutation probability from $(0,0;\ell)$ to $(1,*;*)$ is $(1-\frac{1}{n})\frac{1}{n}$ (where $*$ represents an arbitrary value). Since Item 1 has the largest value-to-weight ratio, after greedy repair,  
only Item 1 remains. The solution becomes $(1,0;0)$. Thus, we get
$$r_{(0,0;\ell), (1,0;0)} \ge p_{(0,0;\ell), (1,0;0)} =\Omega\left(\frac{1}{n}\right).$$ 

Then the lower bound coefficient
\begin{align}
    c_{(0,0;0), (1,0;0)} =  \Omega\left(\frac{1}{n}\right) +  \Omega\left(\frac{1}{n}\right)\sum^{n/2-1}_{\ell=1} c_{(0,0;0), (0,0;\ell)} .
    \label{instance2-b5}
\end{align} 

The lower bound coefficient $c_{(0,0;0), (0,0;\ell)}$ is calculated using Corollary \ref{cor:LowerExplicitExpression}. By \eqref{equ:LowerExplicitExpression} in Corollary \ref{cor:LowerExplicitExpression}, we assign 
\begin{align}
    & c_{(0,0;0), (0,0;\ell)}  = \prod^{\ell-1}_{j=0} {r_{(0,0;j) ,  (0,0;(j, \ell])}} =\prod^{\ell-1}_{j=0} \frac{p_{(0,0;j) ,  (0,0;(j, \ell])}} {p_{(0,0;j) ,  (0,0;j)^+}} \nonumber\\
    &=   \prod^{\ell-1}_{j=0} \frac{p_{(0,0;j) ,  (0,0;(j, \ell])}}{ p_{(0,0;j) ,  (0,0;(j, n-2])} +p_{(0,0;j) ,  (1,0;0)}+ p_{(0,0;j) ,  (0,1;[0,1])}}\nonumber\\
   &= \prod^{\ell-1}_{j=0} \frac{1}{ 1+\frac{p_{(0,0;j), (0,0;[\ell+1,n-2])} + p_{(0,0;j), (1,0;0)}+p_{(0,0;j), (0,1;[0,1])}}{p_{(0,0;j), (0,0;(j, \ell])}}}.
\label{instance2-b6}
\end{align}   
For $j\le \ell-1< n/2$, the transition probabilities  
\begin{align*}
  &  p_{(0,0;j), (0, 0; [\ell+1,n-2])} \le \binom{n-2-j}{\ell+1-j} \left(\frac{1}{n}\right)^{\ell+1-j},\\
   &  p_{(0, 0;j), (0, 1; [0,1])}   \le 
     \frac{1}{n}.
     \end{align*}
\begin{align*}
    &      p_{(0, 0;j), (1, 0; 0)}   \le 
     \frac{1}{n} , 
     \\
      &   p_{(0, 0;j), (0,0;(j,\ell])}  \ge  \binom{n-2-j}{1} \frac{1}{n} \left(1-\frac{1}{n}\right)^{n-1}   . 
\end{align*}
Substituting them into \eqref{instance2-b6}, we get for $\ell<n/2$,
\begin{align*} 
c_{(0,0;0), (0,0;\ell)} &\ge \prod^{\ell-1}_{j=0} \frac{1}{1+ \frac{e}{(\ell+1-j)!} + \frac{2e}{n-2-j} } \\
&\ge \prod^{n/2-1}_{j=0} \frac{1}{1+ \frac{e}{(n/2-j)!} + \frac{2e}{n-2-j} }=\Omega(1).
\end{align*}
We omit the proof of $\Omega(1)$ in the product, which one can refer to \cite[Lemma 2]{he2023fast} for details.  
Then we get
\begin{align*}
    c_{(0,0;0), (1,0;0)} =  \Omega\left(\frac{1}{n}\right) + \Omega\left(\frac{1}{n}\right)\sum^{n/2-1}_{\ell=1} \Omega(1) =\Omega(1).
\end{align*} 

Thus, the mean hitting time to  the global optimal set $L_{(0,0;n-2)}$ is
\begin{align} 
         m_{(0,0;0),(0,0;n-2)} \ge  \frac{\Omega(1) }{(n-1) n^{-n}}.
      \label{instance2-b11}
\end{align} 

By comparing \eqref{instance2-a5} and \eqref{instance2-b11}, we observe that for KP2, the (1+1) EA using greedy repair is slower than that of the (1+1) EA using feasibility rules by a factor $O(n^{-1})$.

\subsection{Instance KP3}
\begin{figure}[ht] 
        \centering
\includegraphics[width=0.4\textwidth]{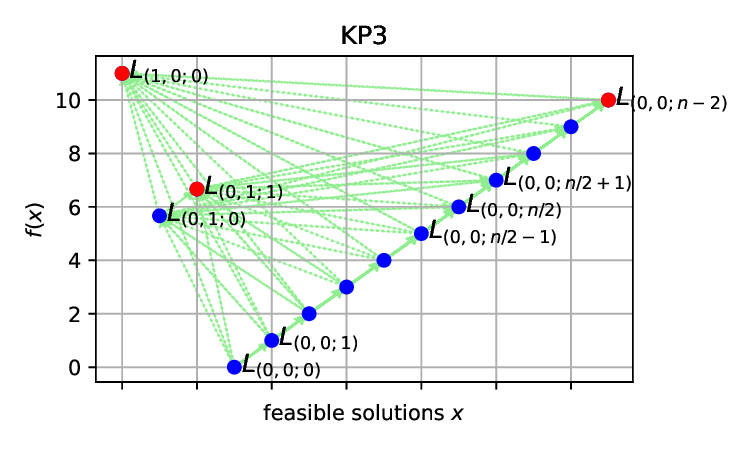} 
        \caption{ The digraph of the two (1+1) EAs on Instance KP3.  Vertices represent fitness levels (feasible solution area). Arcs represent transitions.  
        $n=12$.}
    \label{fig:Knapsack3}
\end{figure}

Fig.~\ref{fig:Knapsack3} shows the digraph of the two (1+1) EAs on Instance KP3.

\subsubsection{The (1+1) EA Using Feasibility Rules} 

According to the lower bound \eqref{equ:LinearLowerBound2} in Theorem \ref{the:TimeBounds}, the mean hitting time from the empty knapsack to the global optimum  $L_{(1,0;0)}$ is
\begin{align}
    m_{(0,0;0), (1,0;0)} \ge     \frac{h_{(0,0;0), (0,0;n-2)}}{p_{(0,0;n-2), (0,0;n-2)^+}}
\label{instance3-a1} . 
\end{align}

Since $L^+_{(0,0;n-2)} = L_{(1,0;0)}$,
 the transition probability  
\begin{align*}
   p_{(0,0;n-2), (1,0;0)}= \left(1-\frac{1}{n}\right)   \left(\frac{1}{n}\right)^{n-1}.
\end{align*}

Thus, the mean hitting time 
\begin{align} 
&  m_{(0,0;0), (1,0;0)} \ge     \Omega(n^{n-1}) h_{(0,0;0), (0,0;n-2)}.
      \label{instance3-a4}
\end{align}

{Intuitively, the chain follows the path $L_{(0,0;0)} \to L_{(0,0;1)} \to \cdots \to L_{(0,0;n-2)}$ and reaches $L_{(0,0;n-2)}$ with positive probability.}
We use Corollary \ref{cor:LowerExplicitExpression} to prove that the hitting probability $h_{(0,0;0),(0,0;n-2)}=\Omega(1)$. By \eqref{equ:LowerExplicitExpression},   we get
\begin{align*}
    & h_{(0,0;0), (0,0;n-2)} \nonumber\\
    &\ge \prod^{n-3}_{j=0} {r_{(0,0;j) ,  (0,0;(j, n-2])}} = \prod^{n-3}_{j=0} \frac{p_{(0,0;j) ,  (0,0;(j, n-2])}}{p_{(0,0;j) ,  (0,0;j)^+}}\nonumber\\
        &\ge \prod^{n-3}_{j=0} \frac{p_{(0,0;j) ,  (0,0;(j, n-2])}}
        {p_{(0,0;j) ,  (0,0;(j, n-2])}+ p_{(0,0;j), (0,1;[0,1])}+p_{(0,0,j),(1,0;0)}}\nonumber\\
   &= \prod^{n-3}_{j=0} \frac{1}{1+\frac{ p_{(0,0;j), (0,1;0)}+p_{(0,0;j), (0,1;1)}+p_{(0,0,j),(1,0;0)}}{p_{(0,0;j), (0,0;(j, n-2])}}}.
\end{align*}  
The transition probabilities 
\begin{align*}
     &p_{(0,0;j), (0,1;0)}\le 
         \left(\frac{1}{n}\right)^{j+1} . \\
    &p_{(0,0;j), (0,1;1)} \le \left\{\begin{array}{ll}
         \frac{1}{n} , & j\le 1,  \\
        \frac{1}{n}\binom{j}{j-1} \left(\frac{1}{n}\right)^{j} ,& j \ge 2.
    \end{array}\right. \\ 
     &p_{(0, 0;j), (1, 0; 0)} \le  
     \left(\frac{1}{n}\right)^{1+j},
     \\
      &   p_{(0, 0;j), (0,0;[j+1,n-2])} \ge  \binom{n-2-j}{1} \frac{1}{n} \left(1-\frac{1}{n}\right)^{n-1}   . 
\end{align*} 
Then we get
\begin{align*}
&h_{(0,0;0), (0,0;j)}\\
& \ge \frac{1}{(1+\frac{3e}{n-2-j})^2}\prod^{n-3}_{j=2} \frac{1}{1 + \frac{2e}{n^j(n-2-j)}+ \frac{ej}{n^{j}(n-2-j)}} =\Omega(1).
\end{align*}
We omit the proof of the bound $\Omega(1)$ in the product, which one can refer to \cite[Lemma 2]{he2023fast} for details.    
By substituting the above bound $\Omega(1)$ into \eqref{instance3-a4}, we get the mean hitting time to the optimal solution as follows:
\begin{align}
m_{(0,0;0),(1,0;0)} =  \Omega(n^{n-1})
\label{instance3-a5}.
\end{align}
\subsubsection{The (1+1) EA with Greedy Repair}   
Let $S_0= L_{(1,0;0)}$ represent the global optimal set, with $S_1$ containing all other feasible solutions. Following the same analysis applied to the (1+1) EA with greedy repair on Instance KP1, we get that the mean hitting time from an empty knapsack to the global optimal solution is given by
\begin{align}
\label{instance3-b1}
m_{(0,0;0),(1,0;0)}=O(n).
\end{align} 

By comparing \eqref{instance2-a5} and \eqref{instance2-b11}, we observe that for KP3, the (1+1) EA using greedy repair is faster than that of the (1+1) EA using feasibility rules by a factor $\Omega(n^{n-2})$. Table~\ref{tab:Comparision} demonstrates that neither greedy repair nor feasibility rules can dominate the other. 

\begin{table}[ht] 
    \centering
    \caption{ Comparison of Algorithm 1 and Algorithm 2.}
    \label{tab:Comparision}
    \begin{tabular}{ccccc cccc}
    \toprule
           &  KP1 &  KP2 & KP3   \\\midrule 
 $\displaystyle \frac{ \mbox{mean hitting time of Algorithm 1}}{\mbox{mean hitting time of Algorihm 2}}$ &  $\Omega(n)$ & $O(n^{-1})$  & $\Omega({n^{n-2}})$   \\ 
            \bottomrule
    \end{tabular} 
\end{table}  

\section{Conclusions}
\label{sec:Conclusions}
This paper investigates the computation of coefficients in the linear drift function for elitist EAs. First, we provide a new interpretation of the linear bound coefficients, where each coefficient corresponds to a hitting probability at a specific fitness level. This transforms the task of estimating the hitting time into one of estimating the hitting probability. Second, we propose a new drift analysis method for estimating hit probability. This method improves the drift analysis method  with new explicit expressions for estimating the hitting time.

The proposed method can estimate both lower and upper bounds on the hitting time, which is useful for comparing the hitting time of two EAs. To demonstrate this, it is applied to compare two EAs with feasibility rules and greedy repair for solving the knapsack problem. The results show that neither constraint handling technique consistently outperforms the other across various instances. However, in certain special cases, using greedy repair can significantly reduce the hitting time from exponential to polynomial. 

Future research will aim to extend this framework to the analysis of more combinatorial optimization problems, such as vertex cover and maximum satisfiability problems. However, there are inherent limitations to the linear drift function. Specifically, it may not provide tight time bounds for fitness functions that do not follow a level-based structure.  Also, it is not available for non-elitist EAs.  
 
\section{Supplement}
In this supplementary material, we examine the connections between Type-$c$, Type-$c_\ell$, and Type-$c_{k,\ell}$ bounds.

\subsection{Necessary and Sufficient Drift Conditions for Type-$c$ and Type-$c_\ell$ Time Bounds}  
In this section, we demonstrate that Type-\(c\) and Type-\(c_\ell\) time bounds can be reformulated in terms of necessary and sufficient drift conditions.

Given a fitness level partition $(S_0,\ldots,S_K)$, and a level  $ S_\ell$ (where $1 \le \ell \le K$),  we use a {drift function} $c(X_k, S_\ell)$ to approximate the hitting probability $c(X_k, S_\ell)$ from $X_k \in S_k$ to $S_\ell$. It is given as follows.  
\begin{equation}
\label{equ:DriftFunction}
    c(X_k, S_\ell) =\left\{
    \begin{array}{ll}
         0 & \mbox{if }    \ell>k,\\
          1    &  \mbox{if }   \ell=k,\\
           1&  \mbox{if }   \ell < k,
    \end{array}
    \right.
\end{equation}   When the coefficient $c_{k,\ell}=c$, the Type-$c_{k,\ell}$ bound  degenerates into the Type-$c$ bound.

\begin{corollary}[Type-$c$ lower bound] 
\label{cor:LowerBound-c} 
Let the coefficient $c_{k,\ell}=c$ in the drift function (\ref{equ:DriftFunction}).  Then  for any $\ell< k $ and $X_k \in S_k$, the drift 
$    {\Delta} c(X_k, S_\ell)  \le 0$   is equivalent to 
\begin{equation} 
\label{equ:LowerCoeff-c} 
c \le \min_{k >\ell} \min_{\ell \ge 1} \; \min_{X_k: p(X_k, S_{[0,\ell]})>0} \frac{p(X_k, S_{\ell})}{p(X_k, S_{[0,\ell]})}.
\end{equation}   
\end{corollary} 

\begin{IEEEproof} It is sufficient to analyze the case $ p(X_k, S_{[0,\ell]})>0$.  
From the drift condition, we have
\begin{align*}
    \begin{array}{rl}
         & {\Delta} c(X_k, S_\ell)  \le 0. \\
         \Leftrightarrow & c \le  r(X_k, S_\ell) + \sum^{k-1}_{j=\ell+1}r(X_{k}, S_j)    c.\\
        \Leftrightarrow &c   \le \frac{r(X_k, S_\ell)}{r(X_k, S_{[0,\ell]}} = \frac{p(X_k, S_{\ell})}{p(X_k, S_{[0,\ell]})}.
    \end{array}
\end{align*} 
Then we obtained the desired conclusion. 
\end{IEEEproof}

Using a similar proof, we can prove the following three corollaries.

\begin{corollary}[Type-$c$ upper bound] 
\label{cor:UpperBound-c} 
Let the coefficient $c_{k,\ell}=c$ in the drift function (\ref{equ:DriftFunction}). Then  for any $k >\ell$ and $X_k \in S_k$, the drift 
$    {\Delta} c(X_k, S_\ell)  \ge 0$  is equivalent to 
\begin{equation} 
\label{equ:UpperCoeff-c} 
c \ge \max_{k>\ell}\; \max_{\ell \ge 1} \; \max_{X_k: p(X_k, S_{[0,\ell]})>0} \frac{p(X_k, S_{\ell})}{p(X_k, S_{[0,\ell]})}.
\end{equation}   
\end{corollary}

As shown in \cite{he2023drift}, the above linear bound coefficient $c$ is the same as  Sudholt's the viscosity in \cite{sudholt2012new}, but is expressed differently.  The above analysis means that $c$ is a lower bound on the hitting probability.

When the coefficient $c_{k,\ell}=c_\ell$, the Type-$c_{k,\ell}$ bound   degenerates into the Type-$c_\ell$ bound. 

\begin{corollary}  [Type-$c_{ \ell}$ lower bound] 
\label{cor:LowerBound-cl} 
Let the coefficient $c_{k,\ell}=c_\ell$ in the drift function (\ref{equ:DriftFunction}). Then  for any $k >\ell$ and $X_k \in S_k$, the drift 
$    {\Delta} c(X_k, S_\ell)  \le 0$  is equivalent to  
\begin{equation} 
\label{equ:LowerCoeff-cl} 
    c_{\ell} \le \min_{k >\ell}  \;   \min_{X_k: p(X_k, S_{[0,\ell]})>0} \frac{p(X_k, S_{\ell})}{p(X_k, S_{[0,\ell]})}.
\end{equation}    
\end{corollary}    

\begin{corollary}  [Type-$c_{ \ell}$ upper bound] 
\label{cor:UpperBound-cl} 
Let the coefficient $c_{k,\ell}=c_\ell$ in the drift function (\ref{equ:DriftFunction}). Then  for any $k >\ell$ and $X_k \in S_k$, the drift 
${\Delta} c(X_k, S_\ell)  \ge 0$  is equivalent to 
\begin{equation} 
\label{equ:UpperCoeff-cl} 
    c_{\ell} \ge \max_{k >\ell}  \;   \max_{X_k: p(X_k, S_{[0,\ell]})>0} \frac{p(X_k, S_{\ell})}{p(X_k, S_{[0,\ell]})}.
\end{equation}    
\end{corollary}

Doerr and K\"otzing~\cite{doerr2024lower} considered random initialization subject to a particular probability distribution \eqref{equ:DK-Condition2}. Adding this condition to Corollary~\ref{cor:LowerBound-cl}, it yields the Type-$c_{ \ell}$ lower bound under random initialization that is identical to \cite[Lemma 10]{doerr2024lower}.

\begin{corollary} 
\label{cor:LowerBound-cl-random} 
If for $\ell \ge 1$, the linear coefficient  $c_{\ell}$ satisfies the two conditions  
\begin{align}
\label{equ:DK-Condition1}
c_\ell &  \le \min_{k >\ell }  \;   \min_{X_k: p(X_k, S_{[0,\ell]})>0} \frac{p(X_k, S_{\ell})}{p(X_k, S_{[0,\ell]})},\\
    \label{equ:DK-Condition2}
c_\ell &\le \frac{\Pr(X^{[0]} \in S_{\ell})}{\Pr(X^{[0]} \in S_{[0,\ell]})}, 
\end{align}  
then the hitting time $$ 
   m(X^{[0]})  \ge \sum^{K}_{\ell=1} \frac{c_{\ell}}{p_{{\scriptscriptstyle\max}}(X_{\ell},S_{[0,\ell-1]})}.$$     
\end{corollary}

\begin{IEEEproof}
For any random initialization,  the hitting time  
\begin{align}
\label{equ:Rand1}
    m(X^{[0]}) =& \sum^K_{\ell=1}  \Pr(X^{[0]} \in S_\ell) \times m(X_\ell).
\end{align}

By Condition \eqref{equ:DK-Condition1} and Corollary \ref{cor:LowerBound-cl}, the hitting time
\begin{align}
\label{equ:Rand2}
   m(X_\ell) \ge   \frac{1 }{p_{{\scriptscriptstyle\max}}(X_{\ell},S_{[0,\ell-1]})}  +  \sum^{\ell-1}_{j=1} \frac{  c_j}{p_{{\scriptscriptstyle\max}}(X_{j},S_{[0,j-1]})}. 
\end{align}
Inserting \eqref{equ:Rand2} to \eqref{equ:Rand1} and reindexing the sum, we get
\begin{align*}
m(X^{[0]}) & \ge \sum^K_{\ell=1} \frac{\Pr(X^{[0]} \in S_\ell)+ c_\ell \sum^{K}_{j=\ell+1}\Pr(X^{[0]} \in S_\ell)}{p_{{\scriptscriptstyle\max}}(X_{\ell},S_{[0,\ell-1]})}.
\end{align*}

Condition \eqref{equ:DK-Condition2} can be rewritten as 
\begin{align}
\label{equ:DK-Condition2-plus}
    c_\ell \le \Pr(X^{[0]} \in S_\ell)+ c_\ell \sum^{K}_{j=\ell+1}\Pr(X^{[0]} \in S_\ell). 
\end{align}

Using \eqref{equ:DK-Condition2-plus}, we get
\begin{align*}
m(X^{[0]}) & \ge   \sum^K_{\ell=1} \frac{ c_\ell}{p_{{\scriptscriptstyle\max}}(X_\ell,S_{[0,\ell-1]})},
\end{align*}
This is the desired result.
\end{IEEEproof}

\paragraph*{Examples (LeadingOnes, OneMax, and Long $k$-path)} 
Consider the (1+1) EA to maximize OneMax, LeadingOnes, and Long $k$-path \cite{sudholt2012new}. The results by Sudholt \cite{sudholt2012new} can be derived by Corollary \ref{cor:LowerBound-c}. The results by Doerr and K\"otzing~\cite{doerr2024lower} can be derived by Corollary \ref{cor:LowerBound-cl-random}.


The above analysis shows that the drift analysis of hitting probability can be used to derive previous results \cite{sudholt2012new,doerr2024lower}. Therefore, its applicability is at least as broad as previous linear bounds.

\subsection{Advantage of the Type-$c_{k,\ell}$ Bound Over the Type-$c_\ell$ Bound}
In most cases, the Type-$c_{k,\ell}$ bound is tighter than the Type-$c$ and Type-$c_{\ell}$ bounds. Let us prove this claim. 
 Since a linear bound is determined by a set of coefficients, the comparison of two linear bounds is based on Pareto dominance.

\begin{definition} Given a fitness level partition $(S_0, \cdots, S_k)$ and an EA, a linear lower bound with coefficients $\{c_{k,\ell} \}$ (where $1 \le \ell < k\le K$)  is considered tighter than another bound with coefficients $\{c'_{k,\ell} \}$  if $c_{k,\ell}  \ge c'_{k,\ell} $ for all $({k,\ell})$, and $c_{k,\ell} > c'_{k,\ell}$ for some $({k,\ell})$. Similarly, a linear upper bound with coefficients $\{c_{k,\ell} \}$  is considered tighter than another bound with coefficients $\{c'_{k,\ell} \}$ if $c_{k,\ell}  \le c'_{k,\ell} $ for all $({k,\ell})$, and $c_{k,\ell} < c'_{k,\ell}$ for some $({k,\ell})$.    
\end{definition}

The theorem below states that the Type-$c_{k,\ell}$ lower bound is tighter than the Type-$c_{\ell}$ lower bound.

\begin{theorem}
\label{the:BestLowerBoundCoeff-cl} Given an elitist and convergent and EA, and  a fitness level partition $(S_0, \cdots, S_K)$, 
For any $1 \le \ell < k$, let $c^*_\ell$ denote the best coefficient $c_{\ell}$ in the Type-$c_\ell$ lower bound such that 
\begin{equation}
\label{equ:BestLowerCoeff-cl}
    c^*_{\ell} = \min_{\ell < i \le k} \; \min_{X_i: p(X_i, S_{[0,\ell]}) > 0} \frac{p(X_i, S_{\ell})}{p(X_i, S_{[0,\ell]})}.
\end{equation}
Let $c^*_{k,\ell}$ denote the coefficient  in the Type-$c_{k,\ell}$ lower bound such that
\begin{align}
\label{equ:BestLowerCoefficient}
    c^*_{k,\ell} = \min_{X_k \in S_k} \left(r(X_{k}, S_\ell) + \sum_{j=\ell+1}^{k-1} r(X_{k}, S_j) \, c^*_{j, \ell}\right).
\end{align}
Then the coefficients $c^*_{k,\ell}$ Pareto dominate $c^*_\ell$, meaning:
\begin{enumerate}
    \item For any $i = \ell + 1, \ldots, k$, $c^*_{i,\ell} \ge c^*_{\ell}$.
    \item If for some $i$,
    \begin{align}
    \label{equ:LowerCoefficient31}
       \min_{X_i: p(X_i, S_{[0,\ell]}) > 0} \frac{p(X_i, S_{\ell})}{p(X_i, S_{[0,\ell]})} > c^*_\ell,
    \end{align}
    then $c^*_{i,\ell} > c^*_{\ell}$.
\end{enumerate} 
\end{theorem}

\begin{IEEEproof}
According to the definition of conditional probability $r(X_k, S_{\ell})$, we get
    \begin{align}\label{equ:LowerCoefficient32}
       \frac{p(X_i, S_{\ell})}{p(X_i, S_{[0,\ell]})} =  \frac{r(X_i, S_{\ell})}{r(X_i, S_{[0,\ell]})}.
    \end{align}
Inserting the above equality to \eqref{equ:BestLowerCoeff-cl}, we get
\begin{align*} 
 c^*_{\ell} = \min_{\ell < i \le k}  \;   \min_{X_i: p(X_i, S_{[0,\ell]})>0}   \frac{r(X_i, S_{\ell})}{r(X_i, S_{[0,\ell]})},  
\end{align*}  
and then for an $i$, 
\begin{align}
\label{equ:LowerCoefficient33}
  r(X_i, S_{\ell}) \ge    r(X_k, S_{[0,\ell]})\; c^*_{\ell}.
\end{align}
Inequality \eqref{equ:LowerCoefficient33} is strict if \eqref{equ:LowerCoefficient31} is true. 
Utilizing Inequality \eqref{equ:LowerCoefficient33}, we can employ induction to demonstrate that $c^*_{k,\ell}$ Pareto dominates $c^*_\ell$. 

Since the conditional probability $r(X_{\ell+1}, S_{[0,\ell]})  =1$, it is trivial to get for $j=\ell+1$,
\begin{align} 
    c^*_{\ell+1,\ell} = r(X_{\ell+1}, S_\ell)=\frac{r(X_{\ell+1}, S_{\ell})}{r(X_{\ell+1}, S_{[0,\ell]})} \ge c^*_{\ell}. 
\end{align} 
The inequality is strict if \eqref{equ:LowerCoefficient31} is true.

Assume that for $j =\ell+1, \cdots, i-1$, $    c^*_{j,\ell}   \ge c^*_{\ell}$. 
Inserting this to \eqref{equ:BestLowerCoefficient}, we have
\begin{align} 
\label{equ:LowerCoefficient34}
    c^*_{i,\ell} \ge  r(X_i, S_\ell)+   \sum^{i-1}_{j=\ell+1} r(X_{i}, S_j)    c^*_{ \ell} .
\end{align}  
Combining \eqref{equ:LowerCoefficient33} with the above inequality, we get
\begin{align} 
\label{equ:LowerCoefficient35}
    c^*_{i,\ell} \ge  r(X_i, S_{[0,\ell]}) \; c^*_{\ell}  +  r(X_i, S_{[\ell+1,i-1]})    c^*_{ \ell} = c^*_\ell.
\end{align}
Inequality \eqref{equ:LowerCoefficient34} is strict if \eqref{equ:LowerCoefficient31} is true. By induction, we complete the proof.  
\end{IEEEproof}

\paragraph*{Examples (OneMax)} 
Even for the (1+1) EA on OneMax, the Type-$c_{k,\ell}$ lower bound is tighter than the Type-$c_{\ell}$ lower bound because the transition probability from level $S_k$ to $S_\ell$ is not equal to the transition probability from $S_i$ to $S_\ell$ if $k \neq i$. 

\paragraph*{Examples (TwoMax1)} 
As proved in \cite{he2023drift}, for the (1+1) EA on TwoMax1, the Type-$c_{k,\ell}$ lower bound is $\Omega(n \ln n)$, but the Type-$c_{\ell}$ is only $O(1)$.  Due to the existence of shortcuts on the multimodal fitness landscapes, Type-$c$ and Type-$c_\ell$ lower bounds are not tight \cite{he2023drift}.

Similarly, the theorem below states that the Type-$c_{k,\ell}$ upper bound is tighter than the Type-$c_{\ell}$ upper bound.

\begin{theorem}
\label{the:BestUpperBoundCoeff-cl} Given an elitist and convergent and EA, and  a fitness level partition $(S_0, \cdots, S_K)$, 
For any $1 \le \ell < k$, let $c^*_\ell$ denote the best coefficient $c_{\ell}$ in the Type-$c_\ell$ upper bound such that 
\begin{equation}
\label{equ:BestUpperCoeff-cl}
    c^*_{\ell} = \max_{\ell < i \le k} \; \max_{X_i: p(X_i, S_{[0,\ell]}) > 0} \frac{p(X_i, S_{\ell})}{p(X_i, S_{[0,\ell]})}.
\end{equation}
Let $c^*_{k,\ell}$ denote the coefficient  in the Type-$c_{k,\ell}$ upper bound such that
\begin{align}
\label{equ:BestupperCoefficient}
    c^*_{k,\ell} = \max_{X_k \in S_k} \left(r(X_{k}, S_\ell) + \sum_{j=\ell+1}^{k-1} r(X_{k}, S_j) \, c^*_{j, \ell}\right).
\end{align}
Then the coefficients $c^*_{k,\ell}$ Pareto dominate $c^*_\ell$, meaning:
\begin{enumerate}
    \item For any $i = \ell + 1, \ldots, k$, $c^*_{i,\ell} \le c^*_{\ell}$.
    \item If for some $i$,
    \begin{align}
    \label{equ:upperrCoefficient31}
       \max_{X_i: p(X_i, S_{[0,\ell]}) > 0} \frac{p(X_i, S_{\ell})}{p(X_i, S_{[0,\ell]})} < c^*_\ell,
    \end{align}
    then $c^*_{i,\ell} <c^*_{\ell}$.
\end{enumerate} 
\end{theorem}

\section*{Acknowledgments}
\noindent S. Y. Chong received research funding under Shenzhen Pengcheng Peacock Special Post.
X. Yao was supported by the National Natural Science Foundation of China (Grant No. 62250710682),  an internal grant from Lingnan University, the Guangdong Provincial Key Laboratory (Grant No. 2020B121201001), and the Program for Guangdong Introducing Innovative and Entrepreneurial Teams (Grant No. 2017ZT07X386). 


\begin{thebibliography}{10}
\providecommand{\url}[1]{#1}
\csname url@samestyle\endcsname
\providecommand{\newblock}{\relax}
\providecommand{\bibinfo}[2]{#2}
\providecommand{\BIBentrySTDinterwordspacing}{\spaceskip=0pt\relax}
\providecommand{\BIBentryALTinterwordstretchfactor}{4}
\providecommand{\BIBentryALTinterwordspacing}{\spaceskip=\fontdimen2\font plus
\BIBentryALTinterwordstretchfactor\fontdimen3\font minus
  \fontdimen4\font\relax}
\providecommand{\BIBforeignlanguage}[2]{{%
\expandafter\ifx\csname l@#1\endcsname\relax
\typeout{** WARNING: IEEEtran.bst: No hyphenation pattern has been}%
\typeout{** loaded for the language `#1'. Using the pattern for}%
\typeout{** the default language instead.}%
\else
\language=\csname l@#1\endcsname
\fi
#2}}
\providecommand{\BIBdecl}{\relax}
\BIBdecl

\bibitem{doerr2021survey}
B.~Doerr and F.~Neumann, ``A survey on recent progress in the theory of
  evolutionary algorithms for discrete optimization,'' \emph{ACM Transactions
  on Evolutionary Learning and Optimization}, vol.~1, no.~4, pp. 1--43, 2021.

\bibitem{doerr2012multiplicative}
B.~Doerr, D.~Johannsen, and C.~Winzen, ``Multiplicative drift analysis,''
  \emph{Algorithmica}, vol.~4, no.~64, pp. 673--697, 2012.

\bibitem{kotzing2019first}
T.~K{\"o}tzing and M.~S. Krejca, ``First-hitting times under drift,''
  \emph{Theoretical Computer Science}, vol. 796, pp. 51--69, 2019.

\bibitem{lengler2020drift}
J.~Lengler, ``Drift analysis,'' in \emph{Theory of Evolutionary Computation:
  Recent Developments in Discrete Optimization}.\hskip 1em plus 0.5em minus
  0.4em\relax Springer, 2020, pp. 89--131.

\bibitem{he2001drift}
J.~He and X.~Yao, ``Drift analysis and average time complexity of evolutionary
  algorithms,'' \emph{Artificial intelligence}, vol. 127, no.~1, pp. 57--85,
  2001.

\bibitem{oliveto2011simplified}
P.~S. Oliveto and C.~Witt, ``Simplified drift analysis for proving lower bounds
  in evolutionary computation,'' \emph{Algorithmica}, vol.~59, no.~3, pp.
  369--386, 2011.

\bibitem{doerr2013adaptive}
B.~Doerr and L.~A. Goldberg, ``Adaptive drift analysis,'' \emph{Algorithmica},
  vol.~65, no.~1, pp. 224--250, 2013.

\bibitem{mitavskiy2009theoretical}
B.~Mitavskiy, J.~Rowe, and C.~Cannings, ``Theoretical analysis of local search
  strategies to optimize network communication subject to preserving the total
  number of links,'' \emph{International Journal of Intelligent Computing and
  Cybernetics}, vol.~2, no.~2, pp. 243--284, 2009.

\bibitem{johannsen2010random}
D.~Johannsen, ``Random combinatorial structures and randomized search
  heuristics,'' Ph.D. dissertation, Universit{\"a}t des Saarlandes
  Saarbr{\"u}cken, 2010.

\bibitem{he2023drift}
J.~He and Y.~Zhou, ``{Drift Analysis with Fitness Levels for Elitist
  Evolutionary Algorithms},'' \emph{Evolutionary Computation}, pp. 1--25, 03
  2024.

\bibitem{wegener2003methods}
I.~Wegener, ``Methods for the analysis of evolutionary algorithms on
  pseudo-boolean functions,'' in \emph{Evolutionary optimization}.\hskip 1em
  plus 0.5em minus 0.4em\relax Springer, 2003, pp. 349--369.

\bibitem{sudholt2012new}
D.~Sudholt, ``A new method for lower bounds on the running time of evolutionary
  algorithms,'' \emph{IEEE Transactions on Evolutionary Computation}, vol.~17,
  no.~3, pp. 418--435, 2012.

\bibitem{doerr2024lower}
B.~Doerr and T.~K{\"o}tzing, ``Lower bounds from fitness levels made easy,''
  \emph{Algorithmica}, vol.~86, no.~2, pp. 367--395, 2024.

\bibitem{yang2023general}
Z.~Yang, H.~Qiu, L.~Gao, D.~Xu, and Y.~Liu, ``A general framework of
  surrogate-assisted evolutionary algorithms for solving computationally
  expensive constrained optimization problems,'' \emph{Information Sciences},
  vol. 619, pp. 491--508, 2023.

\bibitem{molina2024differential}
D.~Molina-P{\'e}rez, E.~Mezura-Montes, E.~A. Portilla-Flores, E.~Vega-Alvarado,
  and B.~Calva-Ya{\~n}ez, ``A differential evolution algorithm for solving
  mixed-integer nonlinear programming problems,'' \emph{Swarm and Evolutionary
  Computation}, vol.~84, p. 101427, 2024.

\bibitem{wang2021set}
R.~Wang and Z.~Zhang, ``Set theory-based operator design in evolutionary
  algorithms for solving knapsack problems,'' \emph{IEEE Transactions on
  Evolutionary Computation}, vol.~25, no.~6, pp. 1133--1147, 2021.

\bibitem{wang2024novel}
L.~Wang, Y.~He, X.~Wang, Z.~Zhou, H.~Ouyang, and S.~Mirjalili, ``A novel
  discrete differential evolution algorithm combining transfer function with
  modulo operation for solving the multiple knapsack problem,''
  \emph{Information Sciences}, vol. 680, p. 121170, 2024.

\bibitem{he2002individual}
J.~He and X.~Yao, ``From an individual to a population: An analysis of the
  first hitting time of population-based evolutionary algorithms,'' \emph{IEEE
  Transactions on Evolutionary Computation}, vol.~6, no.~5, pp. 495--511, 2002.

\bibitem{chen2010choosing}
T.~Chen, J.~He, G.~Chen, and X.~Yao, ``Choosing selection pressure for wide-gap
  problems,'' \emph{Theoretical Computer Science}, vol. 411, no.~6, pp.
  926--934, 2010.

\bibitem{jagerskupper2007algorithmic}
J.~J{\"a}gersk{\"u}pper, ``Algorithmic analysis of a basic evolutionary
  algorithm for continuous optimization,'' \emph{Theoretical Computer Science},
  vol. 379, no.~3, pp. 329--347, 2007.

\bibitem{yuen2006bounds}
S.~Y. Yuen and B.~K.-S. Cheung, ``Bounds for probability of success of
  classical genetic algorithm based on {Hamming} distance,'' \emph{IEEE
  Transactions on Evolutionary Computation}, vol.~10, no.~1, pp. 1--18, 2006.

\bibitem{kotzing2014concentration}
T.~K{\"o}tzing, ``Concentration of first hitting times under additive drift,''
  in \emph{Proceedings of the 2014 Annual Conference on Genetic and
  Evolutionary Computation}, 2014, pp. 1391--1398.

\bibitem{he2003towards}
J.~He and X.~Yao, ``Towards an analytic framework for analysing the computation
  time of evolutionary algorithms,'' \emph{Artificial Intelligence}, vol. 145,
  no. 1-2, pp. 59--97, 2003.

\bibitem{he2016average}
J.~He and G.~Lin, ``Average convergence rate of evolutionary algorithms,''
  \emph{IEEE Transactions on Evolutionary Computation}, vol.~20, no.~2, pp.
  316--321, 2016.

\bibitem{chong2019new}
S.~Y. Chong, P.~Ti{\v{n}}o, J.~He, and X.~Yao, ``A new framework for analysis
  of coevolutionary systems—directed graph representation and random walks,''
  \emph{Evolutionary Computation}, vol.~27, no.~2, pp. 195--228, 2019.

\bibitem{chong2019coevolutionary}
S.~Y. Chong, P.~Ti{\v{n}}o, and J.~He, ``Coevolutionary systems and pagerank,''
  \emph{Artificial Intelligence}, vol. 277, p. 103164, 2019.

\bibitem{he2023fast}
J.~He, S.~Y. Chong, and X.~Yao, ``Fast estimations of hitting time of elitist
  evolutionary algorithms from fitness levels,'' \emph{arXiv}, 2023,
  doi.org/10.48550/arXiv.2311.10502.

\bibitem{norris1998markov}
J.~Norris, \emph{Markov Chains}.\hskip 1em plus 0.5em minus 0.4em\relax
  Cambridge: Cambridge University Press, 1997.

\bibitem{cormen2022introduction}
T.~H. Cormen, C.~E. Leiserson, R.~L. Rivest, and C.~Stein, \emph{Introduction
  to algorithms}.\hskip 1em plus 0.5em minus 0.4em\relax MIT press, 2022.

\bibitem{he2017average}
J.~He and X.~Yao, ``Average drift analysis and population scalability,''
  \emph{IEEE Transactions on Evolutionary Computation}, vol.~21, no.~3, pp.
  426--439, 2017.

\bibitem{mahrach2020comparison}
M.~Mahrach, G.~Miranda, C.~Le{\'o}n, and E.~Segredo, ``Comparison between
  single and multi-objective evolutionary algorithms to solve the knapsack
  problem and the travelling salesman problem,'' \emph{Mathematics}, vol.~8,
  no.~11, p. 2018, 2020.

\bibitem{michalewicz2014genetic}
Z.~Michalewicz, \emph{Genetic Algorithms + Data Structures = Evolution
  Programs}.\hskip 1em plus 0.5em minus 0.4em\relax Springer, 1996.

\bibitem{zhou2007runtime}
Y.~Zhou and J.~He, ``A runtime analysis of evolutionary algorithms for
  constrained optimization problems,'' \emph{IEEE Transactions on Evolutionary
  Computation}, vol.~11, no.~5, pp. 608--619, 2007.

\end{thebibliography}

\end{document}